\documentclass{article}

\usepackage{arxiv}

\usepackage[utf8]{inputenc} \usepackage[T1]{fontenc}    \usepackage{hyperref}       \usepackage{url}            \usepackage{booktabs}       \usepackage{amsfonts}       \usepackage{nicefrac}       \usepackage{microtype}      \usepackage{lipsum}		\usepackage{graphicx}
\graphicspath{./figures}
\usepackage[square,numbers,compress]{natbib}
\usepackage{doi}

\usepackage{amssymb}
\usepackage{amsthm}
\usepackage{amsmath}
\usepackage{caption}
\usepackage{subcaption}
\usepackage{chngcntr}
\usepackage{xcolor}
\usepackage{bm}

\usepackage[capitalize]{cleveref}

\newcommand{\revision}[1]{#1}

\def\eqref#1{equation~\ref{#1}}

\def\1{\bm{1}}

\def\eps{{\epsilon}}

\def\vzero{{\bm{0}}}

\def\vmu{{\bm{\mu}}}
\def\vtheta{{\bm{\theta}}}

\def\vb{{\bm{b}}}

\def\ve{{\bm{e}}}
\def\vf{{\bm{f}}}

\def\vm{{\bm{m}}}

\def\vw{{\bm{w}}}
\def\vx{{\bm{x}}}
\def\vy{{\bm{y}}}
\def\vz{{\bm{z}}}

\def\mH{{\bm{H}}}
\def\mI{{\bm{I}}}
\def\mJ{{\bm{J}}}
\def\mK{{\bm{K}}}

\def\mM{{\bm{M}}}

\def\mU{{\bm{U}}}
\def\mV{{\bm{V}}}

\def\mX{{\bm{X}}}

\def\mZ{{\bm{Z}}}

\def\mSigma{{\bm{\Sigma}}}

\DeclareMathAlphabet{\mathsfit}{\encodingdefault}{\sfdefault}{m}{sl}
\SetMathAlphabet{\mathsfit}{bold}{\encodingdefault}{\sfdefault}{bx}{n}

\def\gC{{\mathcal{C}}}
\def\gD{{\mathcal{D}}}

\def\gI{{\mathcal{I}}}

\def\gN{{\mathcal{N}}}
\def\gO{{\mathcal{O}}}

\def\sI{{\mathbb{I}}}

\newcommand{\E}{\mathbb{E}}
\newcommand{\Ls}{\mathcal{L}}
\newcommand{\R}{\mathbb{R}}

\newcommand{\KL}{D_{\mathrm{KL}}}

\DeclareMathOperator*{\argmax}{arg\,max}
\DeclareMathOperator*{\argmin}{arg\,min}

\newcommand{\params}{\vtheta}
\newcommand{\data}{\gD}
\newcommand{\given}{\, \vert \,}
\newcommand{\diff}{\mathrm{d}}
\newcommand{\hyperparams}{\bm{\psi}}
\newcommand{\variance}{\sigma^2}
\newcommand{\bigO}{\gO}
\newcommand{\vxi}{\bm{\xi}}

\newcommand\annotapprox[1]{\mathrel{\overset{\makebox[0pt]{\mbox{\normalfont\tiny\sffamily #1}}}{\approx}}}

\title{Priors in Bayesian Deep Learning: A Review}

\date{} 					

\author{Vincent Fortuin \\
	Department of Computer Science\\
	ETH Z\"urich\\
	Z\"urich, Switzerland \\
	\texttt{fortuin@inf.ethz.ch}
}

\hypersetup{
pdftitle={Priors in Deep Bayesian Models: A Review},
pdfsubject={stat.ML, cs.LG},
pdfauthor={Vincent~Fortuin},
pdfkeywords={Machine Learning, Bayesian Models, Priors},
}

\begin{document}
\maketitle

\begin{abstract}

While the choice of prior is one of the most critical parts of the Bayesian inference workflow, recent Bayesian deep learning models have often fallen back on vague priors, such as standard Gaussians.
In this review, we highlight the importance of prior choices for Bayesian deep learning and present an overview of different priors that have been proposed for (deep) Gaussian processes, variational autoencoders, and Bayesian neural networks.
We also outline different methods of learning priors for these models from data.
We hope to motivate practitioners in Bayesian deep learning to think more carefully about the prior specification for their models and to provide them with some inspiration in this regard.

\end{abstract}

\section{Introduction}
\label{sec:intro}

Bayesian models have gained a stable popularity in data analysis \citep{gelman2013bayesian} and machine learning \citep{murphy2012machine}.
Especially in recent years, the interest in combining these models with deep learning has surged\footnote{As attested, for instance, by the growing interest in the \href{http://bayesiandeeplearning.org/}{Bayesian Deep Learning} workshop at NeurIPS.}.
The main idea of Bayesian modeling is to infer a \emph{posterior} distribution over the parameters $\params$ of the model given some observed data $\data$ using Bayes' theorem \citep{bayes1763lii, laplace1774memoire} as
\begin{equation}
	\label{eq:posterior}
	p(\params \given \data) = \frac{p(\data \given \params) \, p(\params)}{p(\data)} = \frac{p(\data \given \params) \, p(\params)}{\int p(\data \given \params) \, p(\params) \, \diff \params} 
\end{equation}
where $p(\data | \params)$ is the \emph{likelihood}, $p(\data)$ is the \emph{marginal likelihood} (or evidence), and $p(\params)$ is the \emph{prior}.
The prior can often be parameterized by hyperparameters $\hyperparams$, in which case we will write it as $p(\params; \hyperparams)$ if we want to highlight this dependence.
This posterior can then be used to model new unseen data $\data^*$ using the \emph{posterior predictive}
\begin{equation}
	\label{eq:predictive}
	p(\data^* \given \data) = \int p(\data^* \given \params) \, p(\params \given \data) \, \diff \params 
\end{equation}

The integral in \cref{eq:predictive} is also called the Bayesian model average, because it averages the predictions of all plausible models weighted by their posterior probability.
This is in contrast to standard maximum-likelihood learning, where only one parameter $\params^*$ is used for the predictions as
\begin{equation}
	p(\data^* \given \data) \approx p(\data^* \given \params^*) \quad \text{with} \quad \params^* = \argmax_\params p(\data \given \params) 
\end{equation}
While much previous work has focused on the properties of the posterior predictive \citep{gelman1996posterior, ovadia2019can}, the approximation of the integrals in \cref{eq:posterior} and \cref{eq:predictive} \citep{kass1998markov, wainwright2008graphical, blei2017variational}, or the use of the marginal likelihood for Bayesian model selection \citep{llorente2020marginal, fong2020marginal}, in this thesis we want to shed some light on the often-neglected term in \cref{eq:posterior}: the prior $p(\params)$.

In orthodox Bayesianism, the prior should be chosen in a way such that it accurately reflects our beliefs about the parameters $\params$ before seeing any data \citep{gelman1996bayesian}.
This has been described as being the most crucial part of Bayesian model building, but also the hardest one, since it is often not trivial to map the subjective beliefs of the practitioner unambiguously onto tractable probability distributions \citep{robert2007bayesian}.
However, in practice, choosing the prior is often rather seen as a nuisance, and there have been many attempts to try to avoid having to choose a meaningful prior, for instance, through objective priors \citep{jeffreys1946invariant, jaynes1968prior}, \revision{reference priors \citep{berger1992development},} empirical Bayes \citep{robbins1955empirical}, or combinations of these \citep{klebanov2020objective}.
\revision{One problem with these methods is that in Bayesian deep learning, they are often not tractable due to the high dimensionality of the inference problem, since they either require computing the Fisher information matrix, solving a series of increasingly high-dimensional integrals, or splitting the model parameters into ``parameters of interest'' and ``nuisance parameters'' \citep{berger1992development}.}
Especially in Bayesian deep learning, it is therefore common practice to choose a (seemingly) ``uninformative'' prior, such as a standard Gaussian \citep[c.f.,][]{fortuin2021bayesian}.

This trend is troubling, because choosing a bad prior can have detrimental consequences for the whole inference endeavor.
While the choice of uninformative (or weakly informative) priors is often being motivated by invocation of the asymptotic consistency guarantees of the Bernstein-von-Mises theorem \citep{doob1949application}, this theorem does not in fact hold in many applications, since its regularity conditions are not satisfied \citep{kleijn2012bernstein}.
Moreover, in the non-asymptotic regime of our practical inferences, especially in high-dimensional settings, the prior can have a strong influence on the posterior, often forcing the probability mass onto arbitrary subspaces of the parameter space \citep{gelman2017prior}.
\revision{This means, for instance, that the seemingly innocuous standard Gaussian prior is not uninformative at all \citep{dawid1973marginalization}, but forces the posterior mass onto a thin spherical subspace, which in most cases does not reflect any useful prior knowledge and can severely bias the inference \citep{gelman2006prior, bhadra2016default}.}

Worse yet, prior misspecification can undermine the very properties that compel us to use Bayesian inference in the first place.
For instance, marginal likelihoods can become meaningless under prior misspecification, leading us to choose suboptimal models when using Bayesian model selection \citep{gelman2020holes}.
Moreover, de Finetti's famous Dutch book argument \citep{de1931sul} can be extended to cases where we can be convinced to take wagers that lose money in expectation when using bad priors, which even holds for the aforementioned objective (Jeffreys) priors \citep{eaton2004dutch}.
In a similar vein, Savage's theorem \citep{savage1972foundations}, which promises us optimal decisions under Bayesian decision theory, breaks down under prior misspecification \citep{cerreia2020making}.
Finally, it can even be shown that PAC-Bayesian inference can exceed the Bayesian one in terms of generalization performance when the prior is misspecified \citep{masegosa2019learning, morningstar2020pac}.

On a more optimistic note, the \emph{no-free-lunch} theorem \citep{wolpert1996lack} states that no learning algorithm is universally superior, or in other words, that different learning algorithms outperform each other on different datasets.
Applied to Bayesian learning, this means that there is also no universally preferred prior, but that each task is potentially endowed with its own optimal prior.
Finding (or at least approximating) this optimal prior then offers the potential for significantly improving the performance of the inference or even enabling successful inference in cases where it otherwise would not have been possible.

All these observations should at least motivate us to think a bit more carefully about our priors than is often done in practice.
But do we really have reason to believe that the commonly used priors in Bayesian deep learning are misspecified?
One recent piece of evidence is the fact that in Bayesian linear models, it can be shown that prior misspecification leads to the necessity to temper the posterior for optimal performance (i.e., use a posterior $p_T(\params \given \data) \propto p(\params \given \data)^{1/T}$ for some $T < 1$) \citep{grunwald2017inconsistency}.
And indeed, this need for posterior tempering has also been observed empirically in modern Bayesian deep learning models \citep[e.g.,][]{zhang2018noisy, osawa2019practical, wenzel2020good, fortuin2021bayesian}.

Based on all these insights, it is thus high time that we critically reflect upon our choices of priors in Bayesian deep learning models.
Luckily for us, there are many alternative priors that we could choose over the standard uninformative ones.
This survey shall attempt to provide an overview of them.
We will review existing prior designs for (deep) Gaussian processes in \cref{sec:gp_priors}, for variational autoencoders in \cref{sec:vaes}, and for Bayesian neural networks in \cref{sec:bnns}.
We will then finish by giving some brief outline of methods for learning priors from data in \cref{sec:learning}.

\section{Priors in (Deep) Gaussian Processes}
\label{sec:gp_priors}

Gaussian processes (GPs) have a long history in Bayesian machine learning and enjoy many useful properties \citep{williams1996gaussian, rasmussen2006gaussian}.
They are nonparametric models, which means that we are not actually specifying a prior over parameters $p(\params)$, but instead a prior over functions $p(f)$.
This prior can also have hyperparameters $\hyperparams$, which parameterize a mean function $m_{\hyperparams}$ and a kernel function $k_{\hyperparams}$ as
\begin{equation}
	p(f; \hyperparams) = \mathcal{GP} \left( m_{\hyperparams}(\cdot), k_{\hyperparams}(\cdot, \cdot) \right) 
\end{equation}

This prior is called a Gaussian process because it has the property that when evaluating the function at any finite set of points $\vx$, the function values $\vf := f(\vx)$ are distributed as $p(\vf) = \gN(\vm_\vx, \mK_{\vx\vx})$, where $\vm_\vx = m_{\hyperparams}(\vx)$ is the vector of mean function outputs, the $(i,j)$'th element of the kernel matrix $\mK_{\vx\vx}$ is given by $k_{\hyperparams}(x_i, x_j)$, and the $d$-dimensional multivariate Gaussian $\gN(\vf ; \vmu,\mSigma)$ is
\begin{equation}
\label{eq:gaussian}
	p(\vf) = \gN(\vmu,\mSigma) := \frac{1}{\sqrt{(2\pi)^d \, \det \mSigma}} \exp\left(-\frac{1}{2} (\vf-\vmu)^\top \mSigma^{-1} (\vf-\vmu) \right)
\end{equation}
The Gaussian process can also be seen as an infinite-dimensional version of this multivariate Gaussian distribution, following the Kolmogorov extension theorem \citep{oksendal2013stochastic}.

This model is often combined with a Gaussian observation likelihood $p(\vy \given \vf) = \gN(\vf, \sigma^2 \mI)$, since it then allows for a closed-form posterior inference \citep{rasmussen2006gaussian} on unseen data points $(\vx^*, \vy^*)$ as
\begin{align}
\label{eq:gp_posterior}
	p(\vy^* \given \vx^*, \vx, \vy) &= \gN(\vm^*, \mK^*) \quad \text{with} \\
	\vm^* &= \vm_{\vx^*} + \mK_{\vx^*\vx} \left( \mK_{\vx\vx} + \sigma^2 \mI \right)^{-1} (\vy - \vm_{\vx}) \nonumber \\
	\mK^* &= \mK_{\vx^*\vx^*} - \mK_{\vx^*\vx} \left( \mK_{\vx\vx} + \sigma^2 \mI \right)^{-1} \mK_{\vx\vx^*} + \sigma^2 \mI  \nonumber
\end{align}

While these models are not deep \emph{per se}, there are many ways in which they connect to Bayesian deep learning, which merits their appearance in this thesis.
In the following, we are going to present how GP priors can be parameterized by deep neural networks (\cref{sec:deep_nn_gps}), how GPs can be stacked to build deeper models (\cref{sec:deep_gps}), and how deep neural networks can themselves turn into GPs or be approximated by GPs (\cref{sec:nn_gps}).

\subsection{Gaussian processes parameterized by deep neural networks}
\label{sec:deep_nn_gps}

As mentioned above, the GP prior is determined by the parameterized functions $m_{\hyperparams}$ and $k_{\hyperparams}$.
It will come as no surprise that these functions can be chosen to be deep neural networks (DNNs).
In the case of deep kernels \citep{calandra2016manifold, wilson2016deep, wilson2016stochastic}, however, one has to proceed with care, since most neural network functions will not actually yield proper kernels.
One option to get a kernel out of a neural network is to use the last layer of the network as a feature space and define the kernel to be the inner product in this space, that is, $k_{\hyperparams}(x, x') = \langle \phi(x; \hyperparams), \phi(x'; \hyperparams)\rangle$, where $\phi(\cdot \, ; \hyperparams)$ is the neural network with parameters $\hyperparams$ and $\langle \cdot, \cdot \rangle$ is the inner product.
This actually leads to a Bayesian linear regression in the feature space of the network and is also sometimes called a Bayesian last layer (BLL) model \citep{salakhutdinov2007using, lazaro2010marginalized, snoek2015scalable, ober2020global, kristiadi2020being, watson2021latent}.

Another option to develop deep kernels is to start with a base kernel $k_{\text{base}}(\cdot, \cdot)$, for instance a radial basis function (RBF) kernel $k_{\text{RBF}}(x,x') = \exp(- \lambda (x-x')^2)$ with lengthscale $\lambda$.
This kernel can then be applied in the DNN feature space, yielding the kernel
\begin{equation}
\label{eq:dnn_kernel}
	k_{\hyperparams}(x, x') = k_{\text{base}} (\phi(x; \hyperparams), \phi(x'; \hyperparams))
\end{equation}
If one chooses the linear kernel $k_{\text{lin}}(x,x')=\langle x,x' \rangle$ as the base kernel, this reduces to the BLL model above.
However, when choosing a kernel like the RBF, this model still yields an infinite-dimensional reproducing kernel Hilbert space (RKHS) and thus offers a full GP that does not reduce to a finite Bayesian linear regression.
These approaches can not only lead to very expressive models, but have also been shown to improve properties such as adversarial robustness \citep{bradshaw2017adversarial}.

When using deep mean functions \citep{iwata2017improving, fortuin2019meta} instead of (or in combination with) deep kernels, less precautions have to be taken, since virtually any function is a valid GP mean function.
Thus, the neural network can simply be used as the mean function itself, as $m_{\hyperparams}(x) = \phi(x;\hyperparams)$.
Moreover, deep mean functions in GPs have been related to other popular learning paradigms, such as functional PCA \citep{fortuin2019meta}.
However, the main problem with these, as with the deep kernels above, is the question how to choose them.
Since DNNs are notoriously hard to interpret, choosing their parameters truly \emph{a priori}, that is, before seeing any data, seems like an impossible task.
These approaches are thus usually used in combination with some additional learning algorithms, which set their parameters based on some objective function.
We will provide further details on these algorithms in \cref{sec:learning}.
As an additional sidenote, we would also like to mention a specific type of GP kernel, namely the convolutional kernel \citep{van2017convolutional}, which is not itself parameterized by a neural networks, but inspired by convolutional neural networks (CNNs) in its construction, leading to improved performance on images.

\subsection{Deep Gaussian processes}
\label{sec:deep_gps}

While GPs can be combined with deep neural networks, as we saw in the previous section, they can also be used to construct deep models in their own right.
This is done by adding $k$ additional latent functions $\{ f_1, \dots, f_k \}$ with function outputs $\{ \vf_1, \dots, \vf_k \}$ and latent variables $\{ \vz_1, \dots, \vz_{k-1} \}$, where each function uses the previous latent variable as its inputs, that is, $\vf_{i+1} = f_{i+1}(\vz_i)$ and $\vf_{1} = f_{1}(\vx)$.
In the simplest case, all these latent GPs still have Gaussian latent likelihoods $p(\vz_{i} \given \vf_i) = \gN(\vf_i, \sigma^2_i \mI)$ and a Gaussian output likelihood $p(\vy \given \vf_k) = \gN(\vf_k, \sigma^2_k \mI)$.
If each of these functions is endowed with a GP prior $p(f_i) = \mathcal{GP}(m_{\hyperparams_i}(\cdot), k_{\hyperparams_i}(\cdot,\cdot))$, this model is called a \emph{deep Gaussian process} \citep{damianou2013deep}.
Similarly to deep neural networks, these models can represent increasingly complex distributions with increasing depth, but unlike neural networks, they still offer a fully Bayesian treatment.
Crucially, in contrast to standard GPs, deep GPs can model a larger class of output distributions \citep{duvenaud2014avoiding}, which includes distributions with non-Gaussian marginals \citep{rudner2020inter}.
For increased flexibility, these models can also be coupled with warping functions between the GP layers \citep{dunlop2018deep}.

While these models seem to be strictly superior and preferable to standard GPs, their additional flexibility comes at a price: the posterior inference is not tractable in closed form anymore.
This means that the posterior has to be estimated using approximate inference techniques, such as variational inference \citep{damianou2013deep, salimbeni2019deep}, expectation propagation \citep{bui2016deep}, or amortized inference \citep{dai2016variational}.
A very popular approximate inference technique for GPs is based on so-called inducing points, which are chosen to be a subset of the training points or generally of the training domain \citep{quinonero2005unifying, snelson2007local, titsias2009variational, hensman2013gaussian, fortuin2018scalable}.
This technique can also be extended to inference in deep GPs \citep{damianou2013deep, salimbeni2017doubly} or replaced by variational random features \citep{cutajar2017random}.
Moreover, it has recently been shown that for certain choices of kernels, neural networks can be trained as point estimates for deep GP posteriors \citep{dutordoir2021deep}.

In contrast to the inference techniques, the choice of priors for deep GPs has generally been understudied.
While a deep GP as a whole can model a rather complex prior over functions, the priors for the single layers in terms of $m_{\hyperparams_i}$ and $k_{\hyperparams_i}$ are often chosen to be quite simple, for instance, RBF kernels with different lengthscales \citep{damianou2013deep}.
An exception to this are combinations of deep GPs with the convolutional GP kernels mentioned above, which yield models that are similar in spirit to deep CNNs \citep{kumar2018deep, blomqvist2019deep, dutordoir2020bayesian}.
Moreover, recent software packages for deep GP inference have made it easier to experiment with different priors \citep{tran2018bayesian, dutordoir2021gpflux}.
One can thus be carefully optimistic that research into better deep GP priors will blossom in the years to come.

Another option to build models with more expressive kernels is to actually parameterize the kernel of a GP with \emph{another GP} \citep{tobar2015learning, benton2019function}.
Particularly, the (hierarchical) prior is then $p(f)=\mathcal{GP}(m_{\hyperparams}(\cdot), \hat{k}(\cdot,\cdot))$ with $\hat{k}(x,x') = \mathrm{FT}^{-1}(\exp s(x-x'))$ and $p(s)=\mathcal{GP}(0, k_{\hyperparams}(\cdot,\cdot))$, where $\mathrm{FT}^{-1}$ is the inverse Fourier transform.
This can also be seen as a deep GP with one hidden layer, and it also does not allow closed-form inference, but relies on approximate inference, for instance, using elliptical slice sampling \citep{benton2019function}.
Finally, one can also achieve a similarly expressive model, at lower computational cost, by transforming the GP using a normalizing flow \citep{maronas2021transforming}, which generalizes the idea of a copula process \citep{wilson2010copula}.

\subsection{Gaussian process limits of neural networks}
\label{sec:nn_gps}

Another way to connect GPs to DNNs is via neural network limits.
It has been known for some time now that the function-space prior $p(f)$ induced by a Bayesian neural network (BNN) with a single hidden layer and any independent finite-variance parameter prior $p(\params)$ converges in the limit of infinite width to a GP, due to the central limit theorem \citep{neal1995bayesian, williams1996computing}.
The limiting GP prior is given by
\begin{align}
	p(f) &= \mathcal{GP}(\vzero, k_{\text{NN}}(\cdot,\cdot)) \nonumber \quad \text{with} \\
	k_{\text{NN}}(\vx,\vx') &= \variance_{w_2} \E_{\vw, \vb} \left[ \varphi(\vw^\top \vx + \vb) \, \varphi(\vw^\top \vx' + \vb)  \right] + \variance_{b_2} \quad \text{where} \\
	{\vw} &\sim \gN(\vzero, \variance_{w_1}\mI) \qquad \text{and} \qquad {\vb} \sim \gN(\vzero, \variance_{b_1}\mI) \nonumber
\end{align}
with the prior weight and bias variances $\variance_{w_1}, \variance_{b_1}$ in the first layer, $\variance_{w_2}, \variance_{b_2}$ in the second layer, and nonlinear activation function $\varphi(\cdot)$.
Note that here it is usually assumed that the weight variances are set as $\variance_{w_i} \propto 1/n_i$, where $n_i$ is the number of units in the $i$'th layer.
The kernel $k_{\text{NN}}(\cdot,\cdot)$ is then called the \emph{neural network GP} (NNGP) kernel.
This result has recently been extended to BNNs with ReLU activations \citep{cho2009kernel} and deep BNNs \citep{hazan2015steps, lee2018deep, matthews2018gaussian}, where the lower layer GP kernel takes the same form as above and the kernel for the higher layers assumes the recursive form
\begin{equation}
	k_{\text{NN}}^\ell(\vx,\vx') = \variance_{w_\ell} \E_{(\vz_1, \vz_2) \sim \gN(\vzero, \mK_{\vx\vx'}^{\ell-1})} \left[ \varphi(\vz_1) \, \varphi(\vz_2) \right] + \variance_{b_\ell}
\end{equation}
with $\mK_{\vx\vx'}^{\ell-1}$ being the $2 \times 2$ kernel matrix of the $(\ell-1)$'th layer kernel evaluated at $\vx$ and $\vx'$.
Moreover, these convergence results can also be shown for convolutional BNNs \citep{garriga2018deep, novak2019bayesian}, even with weight correlations \citep{garriga2021correlated}, and for attention neural networks \citep{hron2020infinite}.

While these results only hold for independent finite-variance priors, they can be extended to dependent priors, where they yield GPs that are marginalized over a hyperprior \citep{tsuchida2019richer}, and to infinite-variance priors, where they lead to $\alpha$-stable processes \citep{peluchetti2020stable}.
Excitingly, it has been shown that this convergence of the BNN prior to a stochastic process also implies the convergence of the posterior under mild regularity assumptions \citep{hron2020exact}.
While these results have typically been derived manually, the recent theoretical framework of \emph{tensor programs} allows to rederive them in a unified way, including for recurrent architectures and batch-normalization \citep{yang2019scaling, yang2019tensori, yang2020tensoriii}.
Moreover, it allows to derive limits for networks where only a subset of the layers converge to infinite width, which recovers the models' ability to learn latent features \citep{yang2020feature}.

Not only infinitely wide BNNs can lead to GP limits, but this is also true for infinitely wide standard DNNs.
Crucially however, in this case, the GP arises not as a function-space prior at initialization, but as a model of training under gradient descent \citep{jacot2018neural, lee2020wide}.
Specifically, neural networks under gradient descent training can be shown to follow the kernel gradient of their functional loss with respect to the so-called \emph{neural tangent kernel} (NTK) which is
\begin{equation}
	k_{\text{NTK}}(\vx,\vx') = \mJ_\params(\vx) \mJ_\params(\vx')^\top
\end{equation}
where $\mJ_\params(\vx)$ is the Jacobian of the neural network with respect to the parameters $\params$ evaluated at input $\vx$.
In the limit of infinite width, this kernel becomes stable over training and can be recursively computed as
\begin{align}
	k_{\text{NTK}}^\ell(\vx,\vx') &= k_{\text{NTK}}^{\ell-1}(\vx,\vx') \, \dot{\Sigma}^{\ell}(\vx,\vx') + {\Sigma}^{\ell}(\vx,\vx') \qquad \text{with} \\
	k_{\text{NTK}}^1(\vx,\vx') &= {\Sigma}^{1}(\vx,\vx') = \frac{1}{n_0} \vx^\top \vx' + 1 \nonumber \\
	{\Sigma}^{\ell}(\vx,\vx') &= \E_{(\vz_1, \vz_2) \sim \gN(\vzero, \mK_{\vx\vx'}^{\ell-1})} \left[ \varphi(\vz_1) \, \varphi(\vz_2) \right] \nonumber \\
	\dot{\Sigma}^{\ell}(\vx,\vx') &= \E_{(\vz_1, \vz_2) \sim \gN(\vzero, \mK_{\vx\vx'}^{\ell-1})} \left[ \dot{\varphi}(\vz_1) \, \dot{\varphi}(\vz_2) \right] \nonumber
\end{align}
where $n_0$ is the number of inputs, $\mK_{\vx\vx'}^{\ell-1}$ is again the kernel matrix of the NTK at the previous layer, $\dot{\varphi}(\cdot)$ is the derivative of the activation function, and the ${\Sigma}^{\ell}$ are so-called \emph{activation kernels}.

In the case of finite width, this kernel will not model the training behavior exactly, but there exist approximate corrections \citep{hanin2019finite}.
Interestingly, this same kernel can also be derived from approximate inference in the neural network, leading to an implicit linearization \citep{khan2019approximate}.
This linearization can also be made explicit and can then be used to improve the performance of BNN predictives \citep{immer2021improving} and for fast domain adaptation in multi-task learning \citep{maddox2021fast}.
Moreover, when using the NTK in a kernel machine, such as a support vector machine, it can outperform the original neural network it was derived from, at least in the small-data regime \citep{arora2019harnessing}.
Similarly to the aforementioned NNGP kernels, the NTKs for different architectures can also be rederived using the framework of tensor programs \citep{yang2019scaling, yang2020tensorii} and there exist practical Python packages for the efficient computation of NNGP kernels and NTKs \citep{novak2019neural}.
Finally, it should be noted that this linearization of neural networks has also been linked to the scaling of the parameters and described as \emph{lazy training}, which has been argued to be inferior to standard neural network training \citep{chizat2019lazy}.

\section{Priors in Variational Autoencoders}
\label{sec:vaes}

Moving on from GPs, \revision{another popular class of Bayesian deep learning model is the variational autoencoder (VAE) \citep{kingma2013auto, rezende2014stochastic}.
VAEs are} Bayesian latent variable models which assume a generative process in which the observations $\vx$ are generated from unobserved latent variables $\vz$ through a likelihood $p(\vx \given \vz)$.
In the case of VAEs, this likelihood is parameterized by a neural network which is trained on the observed data.
Since the nonlinearity of this neural network renders exact inference on the posterior $p(\vz \given \vx)$ intractable, it is approximated with a variational approximation $q(\vz \given \vx)$, which is typically also parameterized by a neural network.
The whole model is then trained by optimizing the evidence lower bound (ELBO)
\begin{equation}
\label{eq:vae_elbo}
	\Ls(\vx, \vartheta) = \E_{\vz \sim q_{\vartheta}(\vz \given \vx)} \left[ \log p_{\vartheta}(\vx \given \vz) \right] - \KL(q_{\vartheta}(\vz \given \vx) \, \| \, p(\vz)) \leq p(\vx)
\end{equation}
where $\vartheta$ are the parameters of the likelihood and inference network and $\KL(q \, \| \, p) = \E_q \left[ \log q - \log p \right]$ is the Kullback-Leibler divergence.
In practice, evaluating this term requires taking training examples $\vx$, then computing $q_\vartheta(\vz \given \vx)$ and sampling a $\vz$, and then computing $p_\vartheta(\vx \given \vz)$ and sampling an $\vx$ again.
This is reminiscent of an autoencoder, where $q_\vartheta$ is the encoder and $p_\vartheta$ is the decoder, hence the name of the model.

The likelihood and approximate posterior are usually chosen to be Gaussian.
While the prior is typically also chosen to be standard Gaussian, that is $p(\vz) = \gN(\vzero, \mI)$, there are many other possible choices, which we will explore in the following.
Particularly, we will look at some proper probability distributions that can directly replace the standard Gaussian (\cref{sec:vae_proper}), at some structural priors that also require changes to the architecture (\cref{sec:vae_struct}), and finally at a particularly interesting VAE model with idiosyncratic architecture and prior, namely the neural process (\cref{sec:nps}).

\subsection{Distributional VAE priors}
\label{sec:vae_proper}

We will use the term \emph{distributional} priors to refer to distributions $p(\vz)$ that can be plugged into the standard VAE architecture described above without changing the rest of the model.
However, note that often it can be beneficial to also change the functional form of the variational posterior $q(\vz \given \vx)$ to fit the prior better.
The first type of prior that has been shown to yield some benefits compared to the standard Gaussian one is a spherical prior, namely a von-Mises-Fisher (vMF) prior \citep{davidson2018hyperspherical}, that is
\begin{equation}
\label{eq:von_mises_fisher}
	p(\vz) = \frac{\kappa^{d/2-1}}{(2\pi)^{d/2} \, \gI_{d/2-1}(\kappa)} \exp(\kappa \vmu^\top \vz)
\end{equation}
for a $d$-dimensional latent space, where $\vmu$ is the mean, $\kappa$ is a concentration parameter, and $\gI_k$ is the modified Bessel function of the first kind of order $k$.
This distribution can be seen as a version of the multivariate Gaussian distribution that is supported on the hypersphere.
However, its main disadvantage is that the modified Bessel function cannot generally be computed in closed form and thus has to be approximated numerically.

These hyperspherical priors have been shown to improve VAE performance on benchmark data over standard Gaussian ones, however mostly in low-dimensional latent spaces (up to $d \approx 20$) \citep{davidson2018hyperspherical}.
This could be due to the Gaussian annulus theorem \citep[][Thm.~2.9]{blum2020foundations}, which states that the measure of a multivariate Gaussian in high dimensions concentrates on a hypersphere anyway.
For higher-dimensional latent spaces, it has thus been proposed to replace the vMF prior with a product of lower-dimensional vMF distributions \citep{davidson2019increasing}.

To overcome the numerical issues of the modified Bessel functions, the \emph{power-spherical} distribution has been suggested as a replacement for the vMF \citep{de2020power}.
Its $d$-dimensional density is given by
\begin{equation}
	p(\vz) = \left( 2^{\kappa + d -1} \pi^{(d-1)/2} \frac{\Gamma(\kappa + \frac{d-1}{2})}{\Gamma(\kappa +d -1)} \right)^{-1} \left( 1 + \vmu^\top \vz \right)^\kappa
\end{equation}
where $\vmu$ is again the mean, $\kappa$ the concentration parameter, and $\Gamma(\cdot)$ is the Gamma function.
Since the Gamma function is easier to evaluate than the modified Bessel function, this density allows for closed-form evaluation and reparameterizable sampling.
Empirically, it yields the same performance in VAEs as the vMF prior, while being numerically more stable \citep{de2020power}.

Another type of priors are mixture priors \citep{dilokthanakul2016deep, jiang2017variational, kopf2019mixture}, typically mixtures of Gaussian of the form
\begin{equation}
	p(\vz) = \sum_{i=1}^K \pi_i \, \gN(\vmu_i, \variance_i \mI) \qquad \text{with} \qquad \sum_{i=1}^K \pi_i = 1
\end{equation}
with $K$ mixture components where $\pi_i$ are the mixture weights that are often set to $\pi_i = 1/K$ in the prior.
These priors have been motivated by the idea that the data might consist of clusters, which should also be disjoint in the latent space \citep{dilokthanakul2016deep}, and they have been shown to outperform many other clustering methods on challenging datasets \citep{kopf2019mixture}.
However, similarly to many other clustering methods, one challenge is to choose the number of clusters $K$ a priori.
This can also be optimized automatically, for instance by specifying a stick-breaking or Dirichlet process hyperprior \citep{nalisnick2016stick}, albeit at the cost of more involved inference.

Finally, most of these priors assume independence between data points.
If we have prior knowledge about potential similarity between data points and we can encode it into a kernel function, a Gaussian process can be a powerful prior for a VAE \citep{casale2018gaussian, fortuin2020gp, pearce2020gaussian}.
The prior is usually defined as
\begin{equation}
	p(\mZ) = \gN(\vzero, \mK_{\vz\vz})
\end{equation}
where $\mZ = (\vz_1, \dots, \vz_n)$ is the matrix of latent variables and $\mK_{\vz\vz}$ is again the kernel matrix with $(i,j)$'th element $k(\vz_i, \vz_j)$ for some suitable kernel function $k(\cdot, \cdot)$.
These models have been shown to excel at conditional generation \citep{casale2018gaussian}, time series modeling \citep{pearce2020gaussian}, missing data imputation \citep{fortuin2020gp}, and disentanglement \citep{bhagat2020disentangling, bing2021disentanglement}.
It should be noted that this comes at additional computational cost compared to standard VAEs, since it requires the $\bigO(n^3)$ inversion of the kernel matrix (see \cref{eq:gp_posterior}).
However, this operation can be made more scalable, either through the use of inducing point methods \citep{jazbec2020scalable, ashman2020sparse} (c.f., \cref{sec:deep_gps}) or through factorized kernels \citep{jazbec2020factorized}.
Moreover, depending on the prior knowledge of the generative process, these models can also be extended to use additive GP priors \citep{ramchandran2020longitudinal} or tensor-valued ones \citep{campbell2020tvgp}.

\subsection{Structural VAE priors}
\label{sec:vae_struct}

In contrast to the distributional priors discussed above, we will use the term \emph{structural} priors to refer to priors that do not only change the actual prior distribution $p(\vz)$ in the VAE model, but also the model architecture itself.
Some of these structural priors are extensions of the distributional priors mentioned above.
For instance, the aforementioned Gaussian mixture priors can be extended with a mixture-of-experts decoder, that is, a factorized generative likelihood, where each factor only depends on one of the latent mixture components \citep{kopf2019mixture}.
Another example are the Gaussian process priors, which are defined over the whole latent dataset $\mZ$ and thus benefit from a modified encoder (i.e., inference network), which encodes the complete dataset $\mX$ jointly \citep{fortuin2020gp}.

In addition to these distributional priors with modified architectures, there are also structural priors which could not be realized with the standard VAE architecture.
One example are hierarchical priors \citep{sonderby2016ladder, zhao2017learning, vahdat2020nvae}, such as
\begin{align}
\label{eq:vae_hierarchical}
	p(\vz_1, \dots, \vz_K) &= p(\vz_1) \, \prod_{i=2}^K p(\vz_i \given \vz_{i-1})  \\
	\hspace{-3cm} \text{or} \qquad \qquad p(\vz_1, \dots, \vz_K) &= p(\vz_1) \, \prod_{i=2}^K p(\vz_i \given \vz_{1}, \dots, \vz_{i-1})
	\label{eq:vae_hierarchical2}
\end{align}
We see here that instead of having a single latent variable $\vz$, these models feature $K$ different latent variables $\{ \vz_i, \dots, \vz_K \}$, which depend on each other hierarchically.
These models require additional generative networks to parameterize the conditional probabilities in \cref{eq:vae_hierarchical} or \cref{eq:vae_hierarchical2}, which then enable them to better model data with intrinsically hierarchical features \citep{sonderby2016ladder, zhao2017learning} and to reach state-of-the-art performance in image generation with VAEs \citep{vahdat2020nvae}.

Another type of structural priors are discrete latent priors, such as the VQ-VAE prior \citep{van2017neural}
\begin{equation}
	p(\vz_q) = \frac{1}{|E|} \qquad \text{with}  \qquad \vz_q = \argmin_{\ve \in E} \| \vz_e - \ve \|_2^2
\end{equation}
where $E$ is a finite dictionary of prototypes and $\vz_e$ is a continuous latent variable that is then discretized to $\vz_q$.
Crucially, the prior is not placed over the continuous $\vz_e$, but over the discrete $\vz_q$, namely as a uniform prior over the dictionary $E$.
These discrete latent variables can then be saved very cheaply and thus lead to much stronger compression than standard VAEs \citep{van2017neural}.
When combining these models with the hierarchical latent variables described above, they can also reach competitive image generation performance \citep{razavi2019generating}.
Moreover, these discrete latent variables can be extended to include neighborhood structures such as self-organizing maps \citep{Kohonen1998}, leading to more interpretable latent representations that can also be used for clustering \citep{fortuin2019som, forest2019deep, manduchi2019dpsom}.
\revision{Finally, similar topological priors can also be induced on continuous latent variables using ideas from \emph{persistent homology} \citep{moor2020topological, gabrielsson2020topology}.}

\subsection{Neural processes}
\label{sec:nps}

To conclude this section, we will look at a structural VAE prior that has spawned a lot of interest in recent years and thus deserves its own subsection: the \emph{neural process} (NP).
This model has been independently proposed under the names of \emph{partialVAE} \citep{ma2019eddi} and (conditional) neural process \citep{garnelo2018conditional, garnelo2018neural}, but the latter nomenclature has caught on in the literature.
The main novelty of this VAE architecture is that it not only models the distribution of one type of observed variable $\vx$, but of two variables $(\vx, \vy)$, which can be split into a \emph{context} and \emph{target} set $(\vx, \vy) = (\vx_c, \vy_c) \cap (\vx_t, \vy_t)$.
These sets are conditionally independent given $\vz$, that is, $p(\vx, \vy \given \vz) = p(\vx_c, \vy_c \given \vz) \, p(\vx_t, \vy_t \given \vz)$.
This then allows to infer an unobserved $\vy_t$ based on the other variables using a variational approximation $q(\vz \given \vx_c, \vy_c)$ and the conditional likelihood $p(\vy_t \given \vz, \vx_t)$.
Thus, the model can be used for missing data imputation \citep{ma2019eddi} and regression \citep{garnelo2018neural} tasks.
Note that, since the likelihood is typically conditioned on $\vx_t$ instead of just on $\vz$, this model can be framed as a \emph{conditional VAE} \citep{sohn2015learning}.

One remarkable feature of this model is the used prior, which is namely
\begin{equation}
	p(\vz) = p(\vz \given \vx_c, \vy_c) \approx q(\vz \given \vx_c, \vy_c)
\end{equation}
This means that instead of using an unconditional prior $p(\vz)$ for the full posterior $p(\vz \given \vx, \vy)$, a part of the data (the context set) is used to condition the prior, which is in turn approximated by the variational posterior with reduced conditioning set.
While this is atypical for classical Bayesian inference and generally frowned upon by orthodox Bayesians, it bears resemblence to the data-dependent oracle priors that can be used in PAC-Bayesian bounds and have been shown to make those bounds tighter \citep{rivasplata2020pac, dziugaite2021role}.

The NP model has been heavily inspired by stochastic processes (hence the name) and has been shown to constitute a stochastic process itself under some assumptions \citep{garnelo2018neural}.
Moreover, when the conditional likelihood $p(\vy_t \given \vz, \vx_t)$ is chosen to be an affine transformation, the model is actually equivalent to a Gaussian process with neural network kernel \citep{rudner2018connection}.

Since their inception, NP models have been extended in expressivity in different ways, both in terms of their inference and their generative model.
On the inference side, there are attentive NPs \citep{kim2018attentive}, which endow the encoder with self-attention (and thus make it Turing complete \citep{perez2021attention}), and convolutional (conditional) NPs \citep{gordon2019convolutional, foong2020meta}, which add translation equivariance to the model.
On the generative side, there are functional NPs \citep{louizos2019functional}, which introduce dependence between the predictions by learning a relational graph structure over the latents $\vz$, and Gaussian NPs \citep{bruinsma2021gaussian}, which achieve a similar property by replacing the generative likelihood with a Gaussian process, the mean and kernel of which are inferred based on the latents.

\section{Priors in Bayesian Neural Networks}
\label{sec:bnns}

Bayesian neural networks \citep{mackay1992practical, neal1995bayesian} are neural network models in which the parameters are determined through Bayesian inference (see \cref{eq:posterior}) and predictions are made using the posterior predictive (see \cref{eq:predictive}).
These models have gained increasing popularity in recent years \citep{jospin2020hands}, mostly due to their uncertainty calibration properties \citep{ovadia2019can}.
While many different priors have been proposed for these models \citep[e.g.,][and references therein]{nalisnick2018priors}, it has often been argued that standard Gaussian priors over the parameters are sufficient and that the modeler's inductive biases should be represented through the choice of architecture instead \citep{wilson2020bayesian}.
This view had been supported by preliminary studies on small networks and simple problems that did not find conclusive evidence for the misspecification of Gaussian priors \citep{silvestro2020prior}.

However, in recent work, the adequacy of Gaussian priors has been put into question, particularly by the discovery that Gaussian priors can cause a \emph{cold posterior effect} \citep{wenzel2020good} which is not caused by some other priors \citep{fortuin2021bayesian}.
Following the general considerations regarding prior misspecification (see above), it thus seems advisable to also consider alternative priors for BNNs.
In the following, we will review priors defined in the weight space (\cref{sec:bnn_weight}) and in the function-space (\cref{sec:bnn_function}) and will also show how to extend these ideas to (Bayesian) ensembles of neural networks (\cref{sec:bnn_ensemble}).

\subsection{Weight-space priors}
\label{sec:bnn_weight}

As mentioned before, the most widely used priors for BNNs are isotropic Gaussian (see \cref{eq:gaussian}) priors \citep[e.g.,][]{neal1995bayesian, hernandez2015probabilistic, louizos2017multiplicative, zhang2019cyclical,immer2021improving, dusenberry2020efficient}.
When these priors are used in combination with ReLU nonlinearities, it has been shown that the distributions of activations within the network grow more heavy-tailed with increasing depth \citep{vladimirova2019understanding}.
However, it has also been shown that these networks converge to GPs in the infinite limit (see \cref{sec:nn_gps}), which has famously led Dave MacKay to inquire whether we have ``thrown the baby out with the bath water'' \citep{mackay1998introduction}, since we usually choose BNN models for their increased expressivity over GPs \revision{(where we mean with expressivity the ability to approximate different distributions over function space in their respective predictives)}.
Moreover, Gaussian priors have recently been shown to cause a cold posterior effect in BNNs.
That is, the tempered posterior $p_T(\params \given \data) \propto p(\params \given \data)^{1/T}$ with $T \ll 1$ performs better than the true Bayesian posterior, suggesting prior misspecification \citep{wenzel2020good}.

A simple extension of standard Gaussian priors are matrix-valued Gaussians, which allow for additional correlations between weights \citep{louizos2016structured}.
Their density is given by
\begin{equation}
	p(\params) = \mathcal{MN}(\mM, \mU, \mV) = \frac{\exp \left( - \frac{1}{2} \, \mathrm{tr} \left[ \mV^{-1} (\params - \mM)^{\top} \mU^{-1} (\params - \mM) \right] \right)}{\left[ (2\pi)^p \det U \det V \right]^{\frac{n}{2}}}
\end{equation}
where $\mM$ is the mean matrix, $\mU$ and $\mV$ are the row and column covariances, and $\mathrm{tr}[\cdot]$ is the trace operator.
These matrix-valued Gaussians can then also be used as variational distributions, leading to increased performance compared to isotropic Gaussians on many tasks \citep{louizos2016structured}.

Another way to improve the expressiveness of Gaussian priors is to combine them with hierarchical hyperpriors \citep{graves2011practical, wu2018deterministic}, which has already been proposed in early work on BNNs \citep{mackay1992practical} as
\begin{equation}
\label{eq:hierarchical_prior}
	p(\params) = \int \gN(\vmu, \mSigma) \, p(\mSigma) \, \diff \mSigma
\end{equation}
where $p(\mSigma)$ is a hyperprior over the covariance.
An example of such a hyperprior is the inverse Wishart distribution \citep[e.g.,][]{ober2020global}, which is in $d$ dimensions given by
\begin{equation}
\label{eq:inverse_wishart}
	p(\mSigma) = \mathcal{IW}_d(\nu, \mK) = \frac{ (\det \mK)^{\frac{\nu+d-1}{2}} \, (\det \mSigma)^{-\frac{\nu+2d}{2}} \, \exp \left( -\frac{1}{2} \mathrm{tr} \left[ \mK \mSigma^{-1} \right] \right) }{2^{\frac{(\nu+d-1)d}{2}} \, \Gamma_d(\frac{\nu+d-1}{2}) }
\end{equation}
where $\nu$ are the degrees of freedom and $\mK$ is the mean of $p(\mSigma)$.
When marginalizing the prior in \cref{eq:hierarchical_prior} over the hyperprior in \cref{eq:inverse_wishart}, it turns out that one gets a $d$-dimensional multivariate Student-t distribution with $\nu$ degrees of freedom \citep{shah2014student}, namely
\begin{equation}
	p(\params) = \frac{\Gamma(\frac{\nu+d}{2}) \, (\det \mK)^{-\frac{1}{2}}}{((\nu-2)\pi)^{\frac{n}{2}} \, \Gamma(\frac{\nu}{2})} \left( 1 + \frac{(\params - \vmu)^\top \mK^{-1} (\params - \vmu)}{\nu - 2} \right)^{-\frac{\nu+d}{2}}
\end{equation}
Such distributions have been shown to model the predictive variance more flexibly in stochastic processes \citep{shah2014student} and BNNs \citep{ober2020global}.
Moreover, in BNNs, it has been shown that priors like these, which are heavy-tailed (also including Laplace priors \citep{williams1995bayesian}) and allow for weight correlations, can reduce the cold posterior effect \citep{fortuin2021bayesian}, suggesting that they are less misspecified than isotropic Gaussians.
Finally, when using Student-t priors, it has been shown that one can obtain expressive BNN posteriors even when forcing the posterior mean of the weights to be zero \citep{neklyudov2018variance}, which highlights the flexibility of these distributions.

Another \revision{Gaussian scale mixture prior} is the horseshoe prior \citep{carvalho2009handling}, which is
\begin{equation}
	p(\theta_i) = \gN(0, \tau^2 \sigma_i^2) \quad \text{with} \quad p(\tau) = \gC^+(0, b_0) \quad \text{and} \quad p(\sigma_i) = \gC^+(0, b_1)
\end{equation}
where $b_0$ and $b_1$ are scale parameters and $\gC^+$ is the half-Cauchy distribution
\begin{equation}
\label{eq:half_cauchy}
	p(\sigma) = \gC^+(\mu, b) = \begin{cases}
		\frac{2}{\pi \, b} \left( 1 + \frac{(\sigma - \mu)^2}{b^2} \right)^{-1} & \text{if} \; \sigma \geq \mu \\
		0 & \text{otherwise}
	\end{cases}
\end{equation}
In BNNs, the horseshoe prior can encourage sparsity \citep{ghosh2018structured} and enable interpretable feature selection \citep{overweg2019interpretable}.
It can also be used to aid compression of the neural network weights \citep{louizos2017bayesian}.
Moreover, in application areas such as genomics, where prior knowledge about the signal-to-noise ratio is available, this knowledge can be encoded in such sparsity-inducing hierarchical priors \citep{cui2020informative}.
Interestingly, the popular neural network regularization technique \emph{dropout} \citep{srivastava2014dropout} can also be understood as an approximation to these types of priors \citep{nalisnick2019dropout} \revision{and they can also be used to explicitly model uncertainty over the network architecture, using doubly stochastic inference techniques \citep{hubin2019combining}}.
Finally, Indian buffet process priors can also be used to similarly encourage sparsity and select smaller numbers of weights \citep{kessler2019hierarchical}.

Another interesting prior is the \emph{radial-directional} prior, which disentangles the direction of the weight vector from its length \citep{oh2019radial}.
It is given by
\begin{equation}
	\params = \theta_r \, \params_d \quad \text{with} \quad \theta_r \sim p_\text{rad}(\theta_r) \quad \text{and} \quad \params_d \sim p_\text{dir}(\params_d)
\end{equation}
where $p_{dir}$ is a distribution over the $d$-dimensional unit sphere and $p_\text{rad}$ is a distribution over $\R$.
It has been proposed by \citet{oh2019radial} to use the von-Mises-Fisher distribution (see \cref{eq:von_mises_fisher}) for $p_\text{dir}$ and the half-Cauchy (see \cref{eq:half_cauchy}) for $p_\text{rad}$.
Conversely, \citet{farquhar2020radial} suggest to use a Gaussian for $p_\text{rad}$ and a uniform distribution over the unit sphere for $p_\text{dir}$, which they reparameterize by sampling from a standard Gaussian and normalizing the sampled vectors to unit length.
It should be noted that the idea of the radial-directional prior is related to the \emph{Goldilocks zone} hypothesis, which says that there exists an annulus at a certain distance from the origin which has a particularly high density of high-performing weight vectors \citep{fort2019goldilocks}.

In the specific case of convolutional neural networks for vision tasks, early research has already noted that the weight distributions of the convolutional filters follow the statistics of natural images \citep{srivastava2003advances, simoncelli2009capturing}.
Based on this insight, weight priors have been suggested for Bayesian CNNs that either use correlated Gaussians to encourage weights that are similar for neighboring pixels \citep{fortuin2021bayesian} or Gabor function priors for the whole filters to encourage, for instance, edge detection \citep{pearce2020structured}.

In terms of even more expressive priors, it has been proposed to model the parameters in terms of the units of the neural network instead of the weights themselves \citep{karaletsos2018probabilistic}.
The weight $\theta_{ij}$ between units $i$ and $j$ would then have the prior
\begin{equation}
	p(\theta_{ij}) = g(\vz_i, \vz_j, \eps) \quad \text{with} \quad p(\vz) = p(\eps) = \gN(\vzero, \mI)
\end{equation}
where the function $g$ can be either parameterized by a neural network \citep{karaletsos2018probabilistic} or by a Gaussian process \citep{karaletsos2020hierarchical}.
A similarly implicit model, with even more flexibility, has been proposed by \citet{atanov2018deep} and is simply given by
\begin{equation}
	p(\theta) = g(\vz, \eps) \quad \text{with} \quad p(\vz) = p(\eps) = \gN(\vzero, \mI)
\end{equation}
In both of these priors, the main challenge is to choose the function $g$.
Since this is hard to do manually, the function is usually (meta-)learned (see \cref{sec:learn_bnn}).
Finally, recent work on software packages for BNN inference (e.g., using gradient-guided MCMC inference \citep{garriga2021exact}) has made it easier to try different weight-space priors, thus fostering research to discover better prior distributions \citep{fortuin2021bnnpriors}.

\subsection{Function-space priors}
\label{sec:bnn_function}

As we saw, there are many different weight-space priors that one can choose for Bayesian neural networks.
When using certain non-standard architectures, such as radial basis function networks \citep{lippmann1989pattern}, desired functional properties (e.g., lengthscale or amplitude) can be directly encoded into those priors \citep{coker2019towards}.
However, when using standard BNNs, choosing the right parameter prior can be challenging, since we often have better intuitions about the functions we would expect rather than the parameters themselves.
The trouble is then that the mapping from parameters to functions in neural networks is highly non-trivial due to their many weight-space symmetries \citep{brea2019weight} and complex function-space geometries \citep{fort2019large}.
This has led to an alternative approach to prior specification in BNNs, namely to specify the priors directly in function space, such that
\begin{equation}
	\int \delta(\phi(\cdot \, ; \params)) \, p(\params \given \data) \, \diff \params \approx  p(f \given \data) \propto p(\data \given f) \, p(f)
\end{equation}
where $p(f)$ is the function-space prior, $\phi(\cdot \, ; \params)$ is the function implemented by a neural network with parameters $\params$ and $\delta(\cdot)$ is the Dirac delta measure (in function space).

As we have seen before (c.f., \cref{sec:gp_priors}), Gaussian processes offer an excellent model class to encode functional prior knowledge through the choice of kernel and mean functions, that is, $p(f) = \mathcal{GP}(m(\cdot), k(\cdot, \cdot))$.
It is thus a natural idea to use GP priors as function-space priors for BNNs.
If one applies this idea in the most straightforward way, one can just optimize a posterior that now depends on the KL divergence between the BNN posterior and the GP prior.
However, since this KL is defined in an infinite-dimensional space, it requires approximations, such as Stein kernel gradient estimators \citep{sun2018functional}.
Alternatively, one can first optimize a weight-space distribution on a BNN to minimize the KL divergence with the desired GP prior (e.g., using Monte Carlo estimates) and then use this optimized weight prior as the BNN prior during inference \citep{flam2017mapping}.

While both of these approaches seem reasonable at first sight, it has been discovered that GP and BNN function-space distributions do not actually have the same support and that the true KL divergence is thus infinite (or undefined) \citep{burt2020understanding}.
It has therefore recently been proposed to use the Wasserstein distance instead, although this also requires approximations \citep{tran2020all}.
If one wants to forego the need for a well-defined divergence, one can also use a hypernetwork \citep{ha2016hypernetworks, krueger2017bayesian} as an implicit distribution of BNN weights and then train the network to match the GP samples on a certain set of function outputs \citep{flam2018characterizing}.
Finally, it has recently been discovered that the ridgelet transform \citep{candes1998ridgelets} can be used to approximate GP function-space distributions with BNN weight-space distributions \citep{matsubara2020ridgelet}.
As a sidenote, it should be noted that the reverse can actually be achieved more easily, namely fitting a GP to the outputs of a BNN \citep{ma2019variational}, which can also be of interest in certain applications.

If one does not want to use a GP prior in function space, one can still encode useful functional prior knowledge into BNN priors.
For instance, through the study of the infinite-width limits of BNNs (see \cref{sec:nn_gps}), one finds that the activation function of the network has a strong influence on the functions being implemented and one can, for instance, modulate the smoothness or periodicity of the BNN output by choosing different activation functions \citep{pearce2020expressive}.
Moreover, one can directly define priors over the BNN outputs, which can encode strong prior assumptions about the values that the functions are allowed to take in certain parts of the input space \citep{yang2019output}, that is,
\begin{equation}
	p(\params) = p_{\text{base}}(\params) \, D(\phi(\gC_x; \params), \gC_y) \implies p(\params \given \data) \propto p(\data \given \params) \, D(\phi(\gC_x; \params), \gC_y) \, p_{\text{base}}(\params)
\end{equation}
where $p_{\text{base}}(\params)$ is some base prior in weight space, $(\gC_x, \gC_y)$ are the inputs and outputs in terms of which the functional constraint is defined and $D(\cdot, \cdot)$ is a discrepancy function.
We see that these priors on output constraints end up looking like additional likelihood terms in the posterior and can thus help to encourage specific features of the output function, for instance, to ensure safety features in critical applications.
A similar idea are \emph{noise-contrastive} priors, which are also specified in function space directly through a prior over unseen data $p(\tilde{\data})$ \citep{hafner2020noise}, which yields the prior predictive
\begin{equation}
	p(\data^*) = \iint p(\data^* \given \params) \, p(\params \given \tilde{\data}) \, p(\tilde{\data}) \, \diff \params \, \diff \tilde{\data}
\end{equation}
This prior can encode the belief that the epistemic uncertainty should grow away from the in-distribution data and can thus also lead to more GP-like behavior in BNN posteriors.
Finally, if we have the prior belief that the BNN functions should not be much more complex than the ones of a different function class (e.g., shallower or even linear models), we can use this other class as a functional reference prior and thus regularize the predictive complexity of the model \citep{nalisnick2021predictive}.

\subsection{Bayesian neural network ensembles}
\label{sec:bnn_ensemble}

Deep neural network ensembles, or \emph{deep ensembles}, are a frequentist method similar to the bootstrap \citep{efron1994introduction} that has been used to gain uncertainty estimates in neural networks \citep{lakshminarayanan2017simple}.
However, it has been recently argued that these ensembles actually approximate the BNN posterior predictive \citep{wilson2020bayesian}, that is
\begin{equation}
	p(\data^* \given \data) = \int p(\data^* \given \params) \, p(\params \given \data) \, \diff \params \approx \frac{1}{K} \sum_{i=1}^K p(\data^* \given \params_i)
\end{equation}
where $\params_i$ are the weights of $K$ independently trained ensemble members of the same architecture.
For linear models, ensembles can actually be made to sample \emph{exactly} from the posterior \citep{matthews2017sample}, while in deeper models they can at least provide lower bounds on the marginal likelihood of the true posterior \citep{lyle2020bayesian}.
These models can also be extended to ensembles with different hyperparameters \citep{wenzel2020hyper}, thus also approximating a hierarchical hyperposterior.
Moreover, they can be made more parameter-efficient by sharing certain parameters between ensemble members \citep{wen2019batchensemble}, which can then also be used for approximate BNN inference \citep{dusenberry2020efficient}.
While these models have performed well in many practical tasks \citep{ovadia2019can}, they can still severely overfit in some scenarios \citep{rahaman2020uncertainty}, leading to ill-calibrated uncertainties \citep{yao2019quality}.
However, it has been shown recently that each ensemble member can be combined with a random function that is sampled from a function-space prior \citep{osband2018randomized, osband2019deep}, and that this can indeed yield uncertainties that are conservative with respect to the Bayesian ones \citep{ciosek2019conservative}.
More specifically, the uncertainties of such ensembles are with high probability at least as large as the ones from a Gaussian process with the corresponding NNGP kernel (see \cref{sec:nn_gps}).
These results can also be extended to the NTK \citep{he2020bayesian}.

Another way of making these deep ensembles more Bayesian and incorporating priors are particle-based approximate inference methods, such as Stein variational gradient descent (SVGD) \citep{liu2016stein}.
In SVGD, the ensemble members (or \emph{particles}) are updated according to
\begin{equation}
	\params_i \leftarrow \params_i + \eta \, \phi(\params_i) \quad \text{with} \quad \phi(\params_i) = \sum_{j=1}^K k(\params_i, \params_j) \, \nabla_{\params_j} \log p(\params_j \given \data) - \nabla_{\params_i} k(\params_i, \params_j)
\end{equation}
where $\eta$ is a step-size and $k(\cdot, \cdot)$ is a kernel function in weight space.
With the right step-size schedule, this update rule converges asymptotically to the true posterior \citep{liu2017stein} and even enjoys some non-asymptotic guarantees \citep{korba2020non}.
Moreover, note that it only requires sample-based access to the gradient of the log posterior (and thus also the log prior), which allows it to be used with different weight-space priors \citep{hu2019applying, d2021stein} and even function-space priors, such as GPs \citep{wang2018function}.
Finally, standard deep ensembles can also be directly extended with a kernelized repulsive force, similar to the one in SVGD, which then also leads to asymptotic convergence to the true Bayesian posterior \citep{d2021repulsive}.

\section{(Meta-)Learning Priors}
\label{sec:learning}

So far, we have explored different types of distributions and methods to encode our prior knowledge into Bayesian deep learning models.
But what if we do not have any useful prior knowledge to encode?
While orthodox Bayesianism would prescribe an uninformative prior in such a case \citep{jaynes1968prior, gelman2013bayesian}, there are alternative ways to elicit priors, namely by learning them from data.
If we go the traditional route of Bayesian model selection using the marginal likelihood (the term $p(\data)$ in \cref{eq:posterior}), we can choose a functional form $p(\params ; \hyperparams)$ for the prior and optimize its hyperparameters $\hyperparams$ with respect to this quantity.
This is called \emph{empirical Bayes} \citep{robbins1955empirical} or \emph{type-II maximum likelihood} (ML-II) estimation \citep{rasmussen2006gaussian}.
While there are reasons to be worried about overfitting in such a setting, there are also arguments that the marginal likelihood automatically trades off the goodness of fit with the model complexity and thus leads to model parsimony in the spirit of \emph{Occam's razor} principle \citep{rasmussen2001occam}.

In the case where we have previously solved tasks that are related to the task at hand (so-called \emph{meta-tasks}), we can alternatively also rely on the framework of \emph{learning to learn} \citep{schmidhuber1987evolutionary, thrun1998learning} or \emph{meta-learning} \citep{baxter2000model}.
If we apply this idea to learning priors for Bayesian models in a hierarchical Bayesian way, we arrive at Bayesian meta-learning \citep{heskes1998solving, tenenbaum1999bayesian, fei2003bayesian, lawrence2004learning}.
This can then also be extended to modern gradient-based methods \citep{grant2018recasting, yoon2018bayesian, finn2018probabilistic}.

While these ML-II optimization and Bayesian meta-learning ideas can in principle be used to learn hyperparameters for most of the priors discussed above, we will briefly review some successful examples of their application below.
Following the general structure from above, we will explore learning priors for Gaussian processes (\cref{sec:learn_gp}), variational autoencoders (\cref{sec:learn_vae}), and Bayesian neural networks (\cref{sec:learn_bnn}).

\subsection{Learning GP priors}
\label{sec:learn_gp}

Following the idea of ML-II optimization, we can use the marginal likelihood to select hyperparameters for the mean and kernel functions of GPs.
Conveniently, the marginal likelihood for GPs (with Gaussian observation likelihood) is available in closed form as
\begin{align}
\label{eq:gp_marglik}
	p_{\hyperparams}(\vy \given \vx) &= \int p(\vy \given f, \vx) \, \mathcal{GP}(m_{\hyperparams}(\cdot), k_{\hyperparams}(\cdot, \cdot)) \, \diff f \\
	&= - \frac{1}{2} \left[ (\vy-m(\vx)^\top (\mK_{\vx\vx} + \variance \mI)^{-1} (\vy-m(\vx)) + \log \det (\mK_{\vx\vx} + \variance \mI) + N \log 2 \pi\right] \nonumber
\end{align}
with $N$ being the number of data points, $\mK_{\vx\vx}$ the kernel matrix on the data points, and $\variance$ the noise of the observation likelihood.
We can see that the first term measures the goodness of fit, while the second term (the log determinant of the kernel matrix) measures the complexity of the model and thus incorporates the Occam's razor principle \citep{rasmussen2006gaussian}.

While this quantity can be optimized to select the hyperparameters of simple kernels, such as the lengthscale of an RBF kernel, it can also be used for more expressive ones.
For instance, one can define a \emph{spectral mixture} kernel in the Fourier domain and then optimize the basis functions' coefficients using the marginal likelihood, which can recover a range of different kernel functions \citep{wilson2013gaussian}.
To make the kernels even more expressive, we can also allow for addition and multiplication of different kernels \citep{duvenaud2013structure}, which can ultimately lead to an \emph{automatic statistician} \citep{lloyd2014automatic}, that is, a model that can choose its own problem-dependent kernel combination based on the data and some kernel grammar.
While this model na\"ively scales rather unfavorably due to the size of the combinatorial search space, it can be made more scalable through cheaper approximations \citep{kim2018scaling} or by making the kernel grammar differentiable \citep{sun2018differentiable}.

Another avenue, which was already alluded to above (see \cref{sec:deep_nn_gps}), is to use a neural network to parameterize the kernel.
The first attempt at this trained a deep belief network on the data and then used it as the kernel function \citep{salakhutdinov2007using}, but later approaches optimized the neural network kernel directly using the marginal likelihood \citep{calandra2016manifold}, often in combination with sparse approximations \citep{wilson2016deep} or stochastic variational inference \citep{wilson2016stochastic} for scalability (see \cref{eq:dnn_kernel}).
In this vein, it has recently been proposed to regularize the Lipschitzness of the used neural network, in order for the learned kernel to preserve distances between data points and thus improve its out-of-distribution uncertainties \citep{liu2020simple, fortuin2021deep}.
While all these approaches still rely on the log determinant term in \cref{eq:gp_marglik} to protect them from overfitting, it has been shown that this is unfortunately not effective enough when the employed neural networks are overparameterized \citep{ober2021promises}.
However, this can be remedied by adding a prior over the neural network parameters, thus effectively turning them into BNNs and the whole model into a proper hierarchical Bayesian model.
It should be noted that these techniques cannot only be used to learn GP priors that work well for a particular task, but also to learn certain invariances from data \citep{vdw2018inv} or to fit GP priors to other (implicit) function-space distributions \citep{ma2019variational} (c.f., \cref{sec:bnn_function}).

As mentioned above, if we have related tasks available, we can use them to meta-learn the GP prior.
This can be applied to the kernel \citep{fortuin2019meta, patacchiola2020bayesian} as well as the mean function \citep{fortuin2019meta}, by optimizing the marginal likelihood on these meta-tasks as
\begin{equation}
	\hyperparams^* = \argmax_{\hyperparams} \, \sum_{i=1}^K \log p_{\hyperparams}(\vy_i \given \vx_i) \qquad \text{with} \qquad \data_M = \left\{ (\vx_i, \vy_i) \right\}_{i=1}^K
\end{equation}
where $\data_M$ is the set of meta-tasks.
Note that the mean function can only safely be optimized in this meta-learning setting, but not in the ML-II setting, since \cref{eq:gp_marglik} does not provide any complexity penalty on the mean function and it would thus severely overfit.
While meta-learning does not risk overfitting on the actual training data (since it is not used), it might overfit on the meta-tasks, if there are too few of them, or if they are too similar to each other \citep{qin2018rethink, yin2019meta}.
In the Bayesian meta-learning setting, this can be overcome by specifying a hierarchical hyperprior, which turns out to be equivalent to optimizing a PAC-Bayesian bound \citep{rothfuss2020pacoh}.
This has been shown to successfully meta-learn GP priors from as few as five meta-tasks.

\subsection{Learning VAE priors}
\label{sec:learn_vae}

Variational autoencoders are already trained using the ELBO (see \cref{eq:vae_elbo}), which is a lower bound on the marginal likelihood.
Moreover, their likelihood $p(\vx \given \vz)$ is trained on this objective, as opposed to being fixed a priori as in most other Bayesian models.
One could thus expect that VAEs would be well suited to also learn their prior using their ELBO.
Indeed, the ELBO can be further decomposed as
\begin{equation}
	\Ls(\vx, \vartheta) = \E_{\vz \sim q_{\vartheta}(\vz \given \vx)} \left[ \log p_{\vartheta}(\vx \given \vz) \right] - \sI_{q_{\vartheta}(\vz, \vx)}(\vz, \vx) - \KL(\bar{q}_{\vartheta}(\vz) \, \| \, p(\vz))
\end{equation}
where $\sI_{q_{\vartheta}(\vz, \vx)}(\vz, \vx)$ is the mutual information between $\vz$ and $\vx$ under the joint distribution $q_{\vartheta}(\vz, \vx) = q_{\vartheta}(\vz \given \vx) \, p(\vx)$ and $\bar{q}_{\vartheta}$ is the aggregated approximate posterior $\bar{q}_{\vartheta}(\vz) = \frac{1}{K} \sum_{i=1}^K q_{\vartheta}(\vz \given \vx_i)$.
Since the KL term in this objective is the only term that depends on the prior and the complexity of $q_{\vartheta}(\vz \given \vx)$ is already penalized by the mutual information term, it has been argued that optimizing the prior $p(\vz)$ with respect to the ELBO could be beneficial \citep{hoffman2016elbo}.
One can then show that the optimal prior under this objective is the aggregated posterior $\E_{\vx \sim p(\vx)} \left[ p(\vz \given \vx) \right]$, where $p(\vx)$ is the data distribution \citep{tomczak2018vae}.

As mentioned above, a more expressive family of prior distributions than the common standard Gaussian priors are Gaussian mixture priors \citep{dilokthanakul2016deep} (see \cref{sec:vae_proper}).
In particular, with an increasing number of components, these mixtures can approximate any smooth distribution arbitrarily closely \citep{dalal1983approximating}.
These VAE priors can be optimized using the ELBO \citep{guo2020variational}, however it has been found that this can severely overfit \citep{tomczak2018vae}, highlighting again that the marginal likelihood (or its lower bound) cannot always protect against overfitting (see \cref{sec:learn_gp}).
Instead, it has been proposed to parameterize the mixture components as variational posteriors on certain inducing points, that is
\begin{equation}
	p(\vz) = \frac{1}{K} \sum_{i=1}^K q(\vz \given \vx_i)
\end{equation}
where the $\vx_i$'s are learned \citep{tomczak2018vae}.
This can indeed improve the VAE performance without overfitting, and since the prior is defined in terms of inducing points in data space, it can also straightforwardly be used with hierarchical VAEs \citep{botros2018hierarchical}.

Since mixture models can exacerbate the computation of the KL divergence and require the difficult choice of a number of components $K$, an alternative are implicit priors which are parameterized by learnable functions.
One specific example for image data has been proposed for VAEs in which the latent space preserves the shape of the data, that is, the $\vz$'s are not just vectors, but 2D or 3D tensors.
In such models, one can define a hierarchical prior over $\vz$, which is parameterized by learnable convolutions over the latent dimensions \citep{gulrajani2016pixelvae}.
Another way of specifying a learnable hierarchical prior is to use memory modules, where the prior is then dependent on the stored memories and the memory is learned together with the rest of the model \citep{bornschein2017variational}.
More generally, one can define implicit prior distributions in VAEs as
\begin{equation}
	\vz = g(\vxi ; \hyperparams) \qquad \text{with} \qquad p(\vxi) = \gN(\vzero, \mI)
\end{equation}
where $g(\cdot \, ; \hyperparams)$ is a learnable diffeomorphism, such as a normalizing flow \citep{rezende2015variational}.
This has been successfully demonstrated with RealNVP flows \citep{dinh2016density}, where it has been shown that the VAE can learn very expressive latent representations even with a single latent dimension \citep{huang2017learnable}.
Moreover, it has been shown that using an autoregressive flow \citep{kingma2016improved} in this way for the prior is equivalent to using an inverse autoregressive flow as part of the decoder \citep{chen2016variational}.

Finally, one can also reshape some base prior by a multiplicative term, that is
\begin{equation}
	p(\vz) \propto p_{\text{base}}(\vz) \, \alpha(\vz; \hyperparams) \qquad \text{with} \qquad p_{\text{base}}(\vz) = \gN(\vzero, \mI)
\end{equation}
where $\alpha(\vz; \hyperparams)$ is some learnable acceptance function \citep{bauer2019resampled}.
Depending on the form of the $\alpha$-function, the normalization constant of this prior might be intractable, thus requiring approximations such as accept/reject sampling \citep{bauer2019resampled}.
Interestingly, when defining an energy $E(\vz; \hyperparams) = - \log \alpha(\vz; \hyperparams)$, the model above can be seen as a latent energy-based model \citep{pang2020learning, aneja2020ncp}.
Moreover, when defining this function in terms of a discriminator $d(\cdot)$ in the data space, that is, $\alpha(\vz; \hyperparams) = \E_{\vx \sim p(\vx \given \vz)} \left[ d(\vx; \hyperparams) \right]$, this yields a so-called \emph{pull-back} prior \citep{chen2020vaepp}, which is related to generative adversarial networks \citep{goodfellow2014generative}.

\subsection{Learning BNN priors}
\label{sec:learn_bnn}

Finally, we will consider learning priors for Bayesian neural networks.
Due to the large dimensionality of BNN weight spaces and the complex mapping between weights and functions (see \cref{sec:bnn_weight}), learning BNN priors has not been attempted very often in the literature.
A manual prior specification procedure that may be loosely called ``learning'' is the procedure in \citet{fortuin2021bayesian}, where the authors train standard neural networks using gradient descent and use their empirical weight distributions to inform their prior choices.
When it comes to proper ML-II optimization, BNNs pose an additional challenge, because their marginal likelihoods are typically intractable and even lower bounds are hard to compute.
Learning BNN priors using ML-II has therefore so far only focused on learning the parameters of Gaussian priors in BNNs with Gaussian approximate posteriors, where the posteriors were computed either using moment-matching \citep{wu2018deterministic} or using the Laplace-Generalized-Gauss-Newton method \citep{immer2021scalable}, that is
\begin{align}
	\log p(\data) \; \annotapprox{Lap} \; \log q(\data) \; &\annotapprox{GGN} \; \log p(\data \given \params^*) - \frac{1}{2} \log \det \left(\frac{1}{2 \pi} \hat{\mH}_{\params^*} \right) \\
	\text{with} \qquad \hat{\mH}_{\params^*} &= \mJ_{\params^*}^\top \mH_{\params^*}^L \mJ_{\params^*} + \mH_{\params^*}^P \nonumber
\end{align}
where $q(\data)$ is the marginal likelihood of a Laplace approximation, $\params^* = \argmax_{\params} p(\params \given \data)$ is the maximum a posteriori (MAP) estimate of the parameters, $\hat{\mH}_{\params^*}$ is an approximate Hessian around $\params^*$, $\mJ_{\params^*}$ is the Jacobian of the BNN outputs with respect to the parameters, $\mH_{\params^*}^L$ is the Hessian of the log likelihood, and $\mH_{\params^*}^P$ is the Hessian of the log prior.
Using this approximation, the marginal likelihood is actually differentiable with respect to the prior hyperparameters $\hyperparams$, such that they can be trained together with the BNN posterior \citep{immer2021scalable}.

Again, if meta-tasks are available, one can try to meta-learn the BNN prior.
For CNNs, one can for instance train standard neural networks on the meta-tasks and then learn a generative model (e.g., a VAE) for the filter weights.
This generative model can then be used as a BNN prior for convolutional filters \citep{atanov2018deep}.
In the case of only few meta-tasks, one can also again use PAC-Bayesian bounds to avoid meta-overfitting, at least when meta-learning Gaussian BNN priors \citep{rothfuss2020pacoh}.
Finally, if we do not have access to actual meta-tasks, but we are aware of invariances in our data, we can construct meta-tasks using data augmentation and use them to learn a prior that is (approximately) invariant to these augmentations \citep{nalisnick2018learning}, that is
\begin{equation}
	\hyperparams^* = \argmin_{\hyperparams} \E_{\params \sim p(\params; \hyperparams)} \left[ \E_{\tilde{\vx} \sim q(\tilde{\vx} \given \vx)} \left[ \KL \left( p(\vy \given \vx, \params) \, \| \, p(\vy \given \tilde{\vx}, \params) \right) \right] \right]
\end{equation}
where $q(\tilde{\vx} \given \vx)$ is the data augmentation distribution.

\section{Conclusion}
\label{sec:conclusion}

We have argued that choosing good priors in Bayesian models is crucial to actually achieve the theoretical and empirical properties that they are commonly celebrated for, including uncertainty estimation, model selection, and optimal decision support.
While practitioners in Bayesian deep learning currently often resort to the option of isotropic Gaussian (or similarly uninformative) priors, we have also highlighted that these priors are usually misspecified and can lead to several unintended negative consequences during inference.
On the other hand, well chosen priors can improve performance and even enable novel applications.
Luckily, a plethora of alternative prior choices is available for popular Bayesian deep learning models, such as (deep) Gaussian processes, variational autoencoders, and Bayesian neural networks.
Moreover, in certain cases, useful priors for these models can even be learned from data alone.

We hope that this survey---while necessarily being incomplete in certain ways---has provided the interested reader with a first overview of the existing literature on priors for Bayesian deep learning and with some guidance on how to choose them.
We also hope to encourage practitioners in this field to consider their prior choices a bit more carefully, and to potentially choose one of the priors presented here instead of the standard Gaussian ones, or better yet, to use inspiration from these priors and come up with even better suited ones for their own models.
If only a small fraction of the time usually spent thinking about increasingly elaborate inference techniques will be instead spent on thinking about the priors used, this effort will have been worthwhile.

\subsection*{Acknowledgments}

We acknowledge funding from the Swiss Data Science Center through a PhD fellowship.
We thank Alex Immer, Adri\`a Garriga-Alonso, and Claire Vernade for helpful feedback on the draft and Arnold Weber for constant inspiration.

\bibliographystyle{unsrtnat}
\bibliography{refs/library}

\begin{thebibliography}{265}
\providecommand{\natexlab}[1]{#1}
\providecommand{\url}[1]{\texttt{#1}}
\expandafter\ifx\csname urlstyle\endcsname\relax
  \providecommand{\doi}[1]{doi: #1}\else
  \providecommand{\doi}{doi: \begingroup \urlstyle{rm}\Url}\fi

\bibitem[Gelman et~al.(2013)Gelman, Carlin, Stern, Dunson, Vehtari, and
  Rubin]{gelman2013bayesian}
Andrew Gelman, John~B Carlin, Hal~S Stern, David~B Dunson, Aki Vehtari, and
  Donald~B Rubin.
\newblock \emph{{Bayesian data analysis}}.
\newblock CRC press, 2013.

\bibitem[Murphy(2012)]{murphy2012machine}
Kevin~P Murphy.
\newblock \emph{{Machine learning: a probabilistic perspective}}.
\newblock MIT Press, 2012.

\bibitem[Bayes(1763)]{bayes1763lii}
Thomas Bayes.
\newblock {An essay towards solving a problem in the doctrine of chances.}
\newblock \emph{Philosophical transactions of the Royal Society of London},
  53:\penalty0 370--418, 1763.
\newblock By the late Rev. Mr. Bayes, {FRS} communicated by Mr. Price, in a
  letter to John Canton, {AMFRS}.

\bibitem[Laplace(1774)]{laplace1774memoire}
Pierre~Simon Laplace.
\newblock {M\'emoire sur la probabilit\'e de causes par les \'evenements}.
\newblock \emph{Memoire de l'Academie Royale des Sciences}, 1774.

\bibitem[Gelman et~al.(1996)Gelman, Meng, and Stern]{gelman1996posterior}
Andrew Gelman, Xiao-Li Meng, and Hal Stern.
\newblock {Posterior predictive assessment of model fitness via realized
  discrepancies}.
\newblock \emph{Statistica sinica}, pages 733--760, 1996.

\bibitem[Ovadia et~al.(2019)Ovadia, Fertig, Ren, Nado, Sculley, Nowozin,
  Dillon, Lakshminarayanan, and Snoek]{ovadia2019can}
Yaniv Ovadia, Emily Fertig, Jie Ren, Zachary Nado, David Sculley, Sebastian
  Nowozin, Joshua Dillon, Balaji Lakshminarayanan, and Jasper Snoek.
\newblock {Can you trust your model's uncertainty? Evaluating predictive
  uncertainty under dataset shift}.
\newblock In \emph{Advances in Neural Information Processing Systems}, pages
  13991--14002, 2019.

\bibitem[Kass et~al.(1998)Kass, Carlin, Gelman, and Neal]{kass1998markov}
Robert~E Kass, Bradley~P Carlin, Andrew Gelman, and Radford~M Neal.
\newblock {Markov chain Monte Carlo in practice: a roundtable discussion}.
\newblock \emph{The American Statistician}, 52\penalty0 (2):\penalty0 93--100,
  1998.

\bibitem[Wainwright and Jordan(2008)]{wainwright2008graphical}
Martin~J Wainwright and Michael Jordan.
\newblock {Graphical models, exponential families, and variational inference}.
\newblock \emph{Foundations and Trends{\textregistered} in Machine Learning},
  1\penalty0 (1--2):\penalty0 1--305, 2008.

\bibitem[Blei et~al.(2017)Blei, Kucukelbir, and McAuliffe]{blei2017variational}
David~M Blei, Alp Kucukelbir, and Jon~D McAuliffe.
\newblock {Variational inference: A review for statisticians}.
\newblock \emph{Journal of the American Statistical Association}, 112\penalty0
  (518):\penalty0 859--877, 2017.

\bibitem[Llorente et~al.(2020)Llorente, Martino, Delgado, and
  Lopez-Santiago]{llorente2020marginal}
Fernando Llorente, Luca Martino, David Delgado, and Javier Lopez-Santiago.
\newblock {Marginal likelihood computation for model selection and hypothesis
  testing: an extensive review}.
\newblock \emph{arXiv preprint arXiv:2005.08334}, 2020.

\bibitem[Fong and Holmes(2020)]{fong2020marginal}
Edwin Fong and CC~Holmes.
\newblock {On the marginal likelihood and cross-validation}.
\newblock \emph{Biometrika}, 107\penalty0 (2):\penalty0 489--496, 2020.

\bibitem[Gelman(1996)]{gelman1996bayesian}
Andrew Gelman.
\newblock {Bayesian model-building by pure thought: some principles and
  examples}.
\newblock \emph{Statistica Sinica}, pages 215--232, 1996.

\bibitem[Robert(2007)]{robert2007bayesian}
Christian Robert.
\newblock \emph{{The Bayesian choice: from decision-theoretic foundations to
  computational implementation}}.
\newblock Springer Science \& Business Media, 2007.

\bibitem[Jeffreys(1946)]{jeffreys1946invariant}
Harold Jeffreys.
\newblock {An invariant form for the prior probability in estimation problems}.
\newblock \emph{Proceedings of the Royal Society of London. Series A.
  Mathematical and Physical Sciences}, 186\penalty0 (1007):\penalty0 453--461,
  1946.

\bibitem[Jaynes(1968)]{jaynes1968prior}
Edwin~T Jaynes.
\newblock {Prior probabilities}.
\newblock \emph{IEEE Transactions on systems science and cybernetics},
  4\penalty0 (3):\penalty0 227--241, 1968.

\bibitem[Berger and Bernardo(1992)]{berger1992development}
James~O Berger and Jos{\'e}~M Bernardo.
\newblock {On the development of the reference prior method}.
\newblock \emph{Bayesian statistics}, 4\penalty0 (4):\penalty0 35--60, 1992.

\bibitem[Robbins(1955)]{robbins1955empirical}
Herbert Robbins.
\newblock \emph{{An Empirical Bayes Approach to Statistics}}.
\newblock Office of Scientific Research, US Air Force, 1955.

\bibitem[Klebanov et~al.(2020)Klebanov, Sikorski, Sch{\"u}tte, and
  R{\"o}blitz]{klebanov2020objective}
Ilja Klebanov, Alexander Sikorski, Christof Sch{\"u}tte, and Susanna
  R{\"o}blitz.
\newblock {Objective priors in the empirical Bayes framework}.
\newblock \emph{Scandinavian Journal of Statistics}, 2020.

\bibitem[Fortuin et~al.(2021{\natexlab{a}})Fortuin, Garriga-Alonso, Wenzel,
  R{\"a}tsch, Turner, {van der Wilk}, and Aitchison]{fortuin2021bayesian}
Vincent Fortuin, Adri{\`a} Garriga-Alonso, Florian Wenzel, Gunnar R{\"a}tsch,
  Richard Turner, Mark {van der Wilk}, and Laurence Aitchison.
\newblock {Bayesian Neural Network Priors Revisited}.
\newblock \emph{arXiv preprint arXiv:2102.06571}, 2021{\natexlab{a}}.

\bibitem[Doob(1949)]{doob1949application}
Joseph~L Doob.
\newblock {Application of the theory of martingales}.
\newblock \emph{Le calcul des probabilites et ses applications}, pages 23--27,
  1949.

\bibitem[Kleijn et~al.(2012)Kleijn, van~der Vaart, et~al.]{kleijn2012bernstein}
Bas~JK Kleijn, Aad~W van~der Vaart, et~al.
\newblock {The Bernstein-von-Mises theorem under misspecification}.
\newblock \emph{Electronic Journal of Statistics}, 6:\penalty0 354--381, 2012.

\bibitem[Gelman et~al.(2017)Gelman, Simpson, and Betancourt]{gelman2017prior}
Andrew Gelman, Daniel Simpson, and Michael Betancourt.
\newblock {The prior can often only be understood in the context of the
  likelihood}.
\newblock \emph{Entropy}, 19\penalty0 (10):\penalty0 555, 2017.

\bibitem[Dawid et~al.(1973)Dawid, Stone, and Zidek]{dawid1973marginalization}
A~Philip Dawid, Mervyn Stone, and James~V Zidek.
\newblock {Marginalization paradoxes in Bayesian and structural inference}.
\newblock \emph{Journal of the Royal Statistical Society: Series B
  (Methodological)}, 35\penalty0 (2):\penalty0 189--213, 1973.

\bibitem[Gelman(2006)]{gelman2006prior}
Andrew Gelman.
\newblock {Prior distributions for variance parameters in hierarchical models
  (comment on article by Browne and Draper)}.
\newblock \emph{Bayesian analysis}, 1\penalty0 (3):\penalty0 515--534, 2006.

\bibitem[Bhadra et~al.(2016)Bhadra, Datta, Polson, and
  Willard]{bhadra2016default}
Anindya Bhadra, Jyotishka Datta, Nicholas~G Polson, and Brandon Willard.
\newblock {Default Bayesian analysis with global-local shrinkage priors}.
\newblock \emph{Biometrika}, 103\penalty0 (4):\penalty0 955--969, 2016.

\bibitem[Gelman and Yao(2020)]{gelman2020holes}
Andrew Gelman and Yuling Yao.
\newblock {Holes in Bayesian statistics}.
\newblock \emph{Journal of Physics G: Nuclear and Particle Physics},
  48\penalty0 (1):\penalty0 014002, 2020.

\bibitem[De~Finetti(1931)]{de1931sul}
Bruno De~Finetti.
\newblock {Sul significato soggettivo della probabilita}.
\newblock \emph{Fundamenta mathematicae}, 17\penalty0 (1):\penalty0 298--329,
  1931.

\bibitem[Eaton and Freedman(2004)]{eaton2004dutch}
Morris~L Eaton and David~A Freedman.
\newblock {Dutch book against some 'objective' priors}.
\newblock \emph{Bernoulli}, 10\penalty0 (5):\penalty0 861--872, 2004.

\bibitem[Savage(1972)]{savage1972foundations}
Leonard~J Savage.
\newblock \emph{{The foundations of statistics}}.
\newblock Courier Corporation, 1972.

\bibitem[Cerreia-Vioglio et~al.(2020)Cerreia-Vioglio, Hansen, Maccheroni, and
  Marinacci]{cerreia2020making}
Simone Cerreia-Vioglio, Lars~Peter Hansen, Fabio Maccheroni, and Massimo
  Marinacci.
\newblock {Making Decisions under Model Misspecification}.
\newblock \emph{arXiv preprint arXiv:2008.01071}, 2020.

\bibitem[Masegosa(2019)]{masegosa2019learning}
Andr{\'e}s~R Masegosa.
\newblock {Learning under model misspecification: Applications to variational
  and ensemble methods}.
\newblock \emph{arXiv preprint arXiv:1912.08335}, 2019.

\bibitem[Morningstar et~al.(2020)Morningstar, Alemi, and
  Dillon]{morningstar2020pac}
Warren~R Morningstar, Alexander~A Alemi, and Joshua~V Dillon.
\newblock $\text{PAC}^m$-{Bayes: Narrowing the Empirical Risk Gap in the
  Misspecified Bayesian Regime}.
\newblock \emph{arXiv preprint arXiv:2010.09629}, 2020.

\bibitem[Wolpert(1996)]{wolpert1996lack}
David~H Wolpert.
\newblock {The lack of a priori distinctions between learning algorithms}.
\newblock \emph{Neural computation}, 8\penalty0 (7):\penalty0 1341--1390, 1996.

\bibitem[Gr{\"u}nwald and Van~Ommen(2017)]{grunwald2017inconsistency}
Peter Gr{\"u}nwald and Thijs Van~Ommen.
\newblock {Inconsistency of Bayesian inference for misspecified linear models,
  and a proposal for repairing it}.
\newblock \emph{{B}ayesian Analysis}, 12\penalty0 (4):\penalty0 1069--1103,
  2017.

\bibitem[Zhang et~al.(2018)Zhang, Sun, Duvenaud, and Grosse]{zhang2018noisy}
Guodong Zhang, Shengyang Sun, David Duvenaud, and Roger Grosse.
\newblock {Noisy natural gradient as variational inference}.
\newblock In \emph{{International Conference on Machine Learning}}, pages
  5852--5861. PMLR, 2018.

\bibitem[Osawa et~al.(2019)Osawa, Swaroop, Khan, Jain, Eschenhagen, Turner, and
  Yokota]{osawa2019practical}
Kazuki Osawa, Siddharth Swaroop, Mohammad Emtiyaz~E Khan, Anirudh Jain, Runa
  Eschenhagen, Richard~E Turner, and Rio Yokota.
\newblock {Practical deep learning with Bayesian principles}.
\newblock In \emph{Advances in neural information processing systems}, pages
  4287--4299, 2019.

\bibitem[Wenzel et~al.(2020{\natexlab{a}})Wenzel, Roth, Veeling,
  {\'S}wi{\,{a}}tkowski, Tran, Mandt, Snoek, Salimans, Jenatton, and
  Nowozin]{wenzel2020good}
Florian Wenzel, Kevin Roth, Bastiaan~S Veeling, Jakub {\'S}wi{\,{a}}tkowski,
  Linh Tran, Stephan Mandt, Jasper Snoek, Tim Salimans, Rodolphe Jenatton, and
  Sebastian Nowozin.
\newblock {How good is the Bayes posterior in deep neural networks really?}
\newblock In \emph{International Conference on Machine Learning},
  2020{\natexlab{a}}.

\bibitem[Williams and Rasmussen(1996)]{williams1996gaussian}
Christopher~KI Williams and Carl~E Rasmussen.
\newblock {Gaussian Processes for Regression}.
\newblock In \emph{{Ninth Annual Conference on Neural Information Processing
  Systems (NIPS 1995)}}, pages 514--520. MIT Press, 1996.

\bibitem[Rasmussen and Williams(2006)]{rasmussen2006gaussian}
Carl~Edward Rasmussen and Christopher~KI Williams.
\newblock \emph{{Gaussian Processes for Machine Learning}}.
\newblock MIT Press, 2006.

\bibitem[Oksendal(2013)]{oksendal2013stochastic}
Bernt Oksendal.
\newblock \emph{{Stochastic differential equations: an introduction with
  applications}}.
\newblock Springer Science \& Business Media, 2013.

\bibitem[Calandra et~al.(2016)Calandra, Peters, Rasmussen, and
  Deisenroth]{calandra2016manifold}
Roberto Calandra, Jan Peters, Carl~Edward Rasmussen, and Marc~Peter Deisenroth.
\newblock {Manifold Gaussian processes for regression}.
\newblock In \emph{{2016 International Joint Conference on Neural Networks
  (IJCNN)}}, pages 3338--3345. IEEE, 2016.

\bibitem[Wilson et~al.(2016{\natexlab{a}})Wilson, Hu, Salakhutdinov, and
  Xing]{wilson2016deep}
Andrew~Gordon Wilson, Zhiting Hu, Ruslan Salakhutdinov, and Eric~P Xing.
\newblock {Deep kernel learning}.
\newblock In \emph{{Artificial intelligence and statistics}}, pages 370--378.
  PMLR, 2016{\natexlab{a}}.

\bibitem[Wilson et~al.(2016{\natexlab{b}})Wilson, Hu, Salakhutdinov, and
  Xing]{wilson2016stochastic}
Andrew~Gordon Wilson, Zhiting Hu, Ruslan Salakhutdinov, and Eric~P Xing.
\newblock {Stochastic variational deep kernel learning}.
\newblock \emph{arXiv preprint arXiv:1611.00336}, 2016{\natexlab{b}}.

\bibitem[Salakhutdinov and Hinton(2007)]{salakhutdinov2007using}
Ruslan Salakhutdinov and Geoffrey Hinton.
\newblock {Using deep belief nets to learn covariance kernels for Gaussian
  processes}.
\newblock In \emph{{Proceedings of the 20th International Conference on Neural
  Information Processing Systems}}, pages 1249--1256, 2007.

\bibitem[L{\'a}zaro-Gredilla and Figueiras-Vidal(2010)]{lazaro2010marginalized}
Miguel L{\'a}zaro-Gredilla and An{\'\i}bal~R Figueiras-Vidal.
\newblock {Marginalized neural network mixtures for large-scale regression}.
\newblock \emph{IEEE transactions on neural networks}, 21\penalty0
  (8):\penalty0 1345--1351, 2010.

\bibitem[Snoek et~al.(2015)Snoek, Rippel, Swersky, Kiros, Satish, Sundaram,
  Patwary, Prabhat, and Adams]{snoek2015scalable}
Jasper Snoek, Oren Rippel, Kevin Swersky, Ryan Kiros, Nadathur Satish,
  Narayanan Sundaram, Mostofa Patwary, Mr~Prabhat, and Ryan Adams.
\newblock {Scalable bayesian optimization using deep neural networks}.
\newblock In \emph{{International conference on machine learning}}, pages
  2171--2180. PMLR, 2015.

\bibitem[Ober and Aitchison(2020)]{ober2020global}
Sebastian~W Ober and Laurence Aitchison.
\newblock {Global inducing point variational posteriors for Bayesian neural
  networks and deep Gaussian processes}.
\newblock \emph{arXiv preprint arXiv:2005.08140}, 2020.

\bibitem[Kristiadi et~al.(2020)Kristiadi, Hein, and Hennig]{kristiadi2020being}
Agustinus Kristiadi, Matthias Hein, and Philipp Hennig.
\newblock {Being bayesian, even just a bit, fixes overconfidence in relu
  networks}.
\newblock In \emph{{International Conference on Machine Learning}}, pages
  5436--5446. PMLR, 2020.

\bibitem[Watson et~al.(2021)Watson, Lin, Klink, Pajarinen, and
  Peters]{watson2021latent}
Joe Watson, Jihao~Andreas Lin, Pascal Klink, Joni Pajarinen, and Jan Peters.
\newblock {Latent Derivative Bayesian Last Layer Networks}.
\newblock In \emph{{International Conference on Artificial Intelligence and
  Statistics}}, pages 1198--1206. PMLR, 2021.

\bibitem[Bradshaw et~al.(2017)Bradshaw, Matthews, and
  Ghahramani]{bradshaw2017adversarial}
John Bradshaw, Alexander G de~G Matthews, and Zoubin Ghahramani.
\newblock {Adversarial examples, uncertainty, and transfer testing robustness
  in Gaussian process hybrid deep networks}.
\newblock \emph{arXiv preprint arXiv:1707.02476}, 2017.

\bibitem[Iwata and Ghahramani(2017)]{iwata2017improving}
Tomoharu Iwata and Zoubin Ghahramani.
\newblock {Improving output uncertainty estimation and generalization in deep
  learning via neural network Gaussian processes}.
\newblock \emph{arXiv preprint arXiv:1707.05922}, 2017.

\bibitem[Fortuin et~al.(2019{\natexlab{a}})Fortuin, Strathmann, and
  R{\"a}tsch]{fortuin2019meta}
Vincent Fortuin, Heiko Strathmann, and Gunnar R{\"a}tsch.
\newblock {Meta-Learning Mean Functions for Gaussian Processes}.
\newblock \emph{arXiv preprint arXiv: 1901.08098}, 2019{\natexlab{a}}.

\bibitem[van~der Wilk et~al.(2017)van~der Wilk, Rasmussen, and
  Hensman]{van2017convolutional}
Mark van~der Wilk, Carl~Edward Rasmussen, and James Hensman.
\newblock {Convolutional Gaussian processes}.
\newblock In \emph{{Proceedings of the 31st International Conference on Neural
  Information Processing Systems}}, pages 2845--2854, 2017.

\bibitem[Damianou and Lawrence(2013)]{damianou2013deep}
Andreas Damianou and Neil~D Lawrence.
\newblock {Deep gaussian processes}.
\newblock In \emph{{Artificial intelligence and statistics}}, pages 207--215.
  PMLR, 2013.

\bibitem[Duvenaud et~al.(2014)Duvenaud, Rippel, Adams, and
  Ghahramani]{duvenaud2014avoiding}
David Duvenaud, Oren Rippel, Ryan Adams, and Zoubin Ghahramani.
\newblock {Avoiding pathologies in very deep networks}.
\newblock In \emph{{Artificial Intelligence and Statistics}}, pages 202--210.
  PMLR, 2014.

\bibitem[Rudner et~al.(2020)Rudner, Sejdinovic, and Gal]{rudner2020inter}
Tim~GJ Rudner, Dino Sejdinovic, and Yarin Gal.
\newblock {Inter-domain deep Gaussian processes}.
\newblock In \emph{{International Conference on Machine Learning}}, pages
  8286--8294. PMLR, 2020.

\bibitem[Dunlop et~al.(2018)Dunlop, Girolami, Stuart, and
  Teckentrup]{dunlop2018deep}
Matthew~M Dunlop, Mark~A Girolami, Andrew~M Stuart, and Aretha~L Teckentrup.
\newblock {How deep are deep Gaussian processes?}
\newblock \emph{Journal of Machine Learning Research}, 19\penalty0
  (54):\penalty0 1--46, 2018.

\bibitem[Salimbeni et~al.(2019)Salimbeni, Dutordoir, Hensman, and
  Deisenroth]{salimbeni2019deep}
Hugh Salimbeni, Vincent Dutordoir, James Hensman, and Marc Deisenroth.
\newblock {Deep Gaussian processes with importance-weighted variational
  inference}.
\newblock In \emph{{International Conference on Machine Learning}}, pages
  5589--5598. PMLR, 2019.

\bibitem[Bui et~al.(2016)Bui, Hern{\'a}ndez-Lobato, Hernandez-Lobato, Li, and
  Turner]{bui2016deep}
Thang Bui, Daniel Hern{\'a}ndez-Lobato, Jose Hernandez-Lobato, Yingzhen Li, and
  Richard Turner.
\newblock {Deep Gaussian processes for regression using approximate expectation
  propagation}.
\newblock In \emph{{International conference on machine learning}}, pages
  1472--1481. PMLR, 2016.

\bibitem[Dai et~al.(2016)Dai, Damianou, Gonz{\'a}lez, and
  Lawrence]{dai2016variational}
Zhenwen Dai, Andreas~C Damianou, Javier Gonz{\'a}lez, and Neil~D Lawrence.
\newblock {Variational Auto-encoded Deep Gaussian Processes.}
\newblock In \emph{{ICLR}}, 2016.

\bibitem[Qui{\~n}onero-Candela and Rasmussen(2005)]{quinonero2005unifying}
Joaquin Qui{\~n}onero-Candela and Carl~Edward Rasmussen.
\newblock {A unifying view of sparse approximate Gaussian process regression}.
\newblock \emph{Journal of Machine Learning Research}, 6\penalty0
  (Dec):\penalty0 1939--1959, 2005.

\bibitem[Snelson and Ghahramani(2007)]{snelson2007local}
Edward Snelson and Zoubin Ghahramani.
\newblock {Local and global sparse Gaussian process approximations}.
\newblock In \emph{{Artificial Intelligence and Statistics}}, pages 524--531.
  PMLR, 2007.

\bibitem[Titsias(2009)]{titsias2009variational}
Michalis Titsias.
\newblock {Variational learning of inducing variables in sparse Gaussian
  processes}.
\newblock In \emph{{Artificial intelligence and statistics}}, pages 567--574.
  PMLR, 2009.

\bibitem[Hensman et~al.(2013)Hensman, Fusi, and Lawrence]{hensman2013gaussian}
James Hensman, Nicol{\`o} Fusi, and Neil~D Lawrence.
\newblock {Gaussian processes for Big data}.
\newblock In \emph{{Proceedings of the Twenty-Ninth Conference on Uncertainty
  in Artificial Intelligence}}, pages 282--290, 2013.

\bibitem[Fortuin et~al.(2021{\natexlab{b}})Fortuin, Dresdner, Strathmann, and
  R{\"a}tsch]{fortuin2018scalable}
Vincent Fortuin, Gideon Dresdner, Heiko Strathmann, and Gunnar R{\"a}tsch.
\newblock {Sparse Gaussian Processes on Discrete Domains}.
\newblock \emph{IEEE Access}, 9, 2021{\natexlab{b}}.

\bibitem[Salimbeni and Deisenroth(2017)]{salimbeni2017doubly}
Hugh Salimbeni and Marc~Peter Deisenroth.
\newblock {Doubly stochastic variational inference for deep Gaussian
  processes}.
\newblock In \emph{{Proceedings of the 31st International Conference on Neural
  Information Processing Systems}}, pages 4591--4602, 2017.

\bibitem[Cutajar et~al.(2017)Cutajar, Bonilla, Michiardi, and
  Filippone]{cutajar2017random}
Kurt Cutajar, Edwin~V Bonilla, Pietro Michiardi, and Maurizio Filippone.
\newblock {Random feature expansions for deep Gaussian processes}.
\newblock In \emph{{International Conference on Machine Learning}}, pages
  884--893. PMLR, 2017.

\bibitem[Dutordoir et~al.(2021{\natexlab{a}})Dutordoir, Hensman, van~der Wilk,
  Ek, Ghahramani, and Durrande]{dutordoir2021deep}
Vincent Dutordoir, James Hensman, Mark van~der Wilk, Carl~Henrik Ek, Zoubin
  Ghahramani, and Nicolas Durrande.
\newblock {Deep Neural Networks as Point Estimates for Deep Gaussian
  Processes}.
\newblock \emph{arXiv preprint arXiv:2105.04504}, 2021{\natexlab{a}}.

\bibitem[Kumar et~al.(2018)Kumar, Singh, Srijith, and Damianou]{kumar2018deep}
Vinayak Kumar, Vaibhav Singh, PK~Srijith, and Andreas Damianou.
\newblock {Deep Gaussian processes with convolutional kernels}.
\newblock \emph{arXiv preprint arXiv:1806.01655}, 2018.

\bibitem[Blomqvist et~al.(2019)Blomqvist, Kaski, and
  Heinonen]{blomqvist2019deep}
Kenneth Blomqvist, Samuel Kaski, and Markus Heinonen.
\newblock {Deep convolutional Gaussian processes}.
\newblock In \emph{{Joint European Conference on Machine Learning and Knowledge
  Discovery in Databases}}, pages 582--597. Springer, 2019.

\bibitem[Dutordoir et~al.(2020)Dutordoir, Wilk, Artemev, and
  Hensman]{dutordoir2020bayesian}
Vincent Dutordoir, Mark Wilk, Artem Artemev, and James Hensman.
\newblock {Bayesian image classification with deep convolutional Gaussian
  processes}.
\newblock In \emph{International Conference on Artificial Intelligence and
  Statistics}, pages 1529--1539. PMLR, 2020.

\bibitem[Tran et~al.(2018)Tran, Dusenberry, van~der Wilk, and
  Hafner]{tran2018bayesian}
Dustin Tran, Michael~W Dusenberry, Mark van~der Wilk, and Danijar Hafner.
\newblock {Bayesian layers: A module for neural network uncertainty}.
\newblock \emph{arXiv preprint arXiv:1812.03973}, 2018.

\bibitem[Dutordoir et~al.(2021{\natexlab{b}})Dutordoir, Salimbeni, Hambro,
  McLeod, Leibfried, Artemev, van~der Wilk, Hensman, Deisenroth, and
  John]{dutordoir2021gpflux}
Vincent Dutordoir, Hugh Salimbeni, Eric Hambro, John McLeod, Felix Leibfried,
  Artem Artemev, Mark van~der Wilk, James Hensman, Marc~P Deisenroth, and
  ST~John.
\newblock {GPflux: A Library for Deep Gaussian Processes}.
\newblock \emph{arXiv preprint arXiv:2104.05674}, 2021{\natexlab{b}}.

\bibitem[Tobar et~al.(2015)Tobar, Bui, and Turner]{tobar2015learning}
Felipe Tobar, Thang~D Bui, and Richard~E Turner.
\newblock {Learning stationary time series using Gaussian processes with
  nonparametric kernels}.
\newblock In \emph{Proceedings of the 28th International Conference on Neural
  Information Processing Systems-Volume 2}, pages 3501--3509, 2015.

\bibitem[Benton et~al.(2019)Benton, Maddox, Salkey, Albinati, and
  Wilson]{benton2019function}
Gregory~W Benton, Wesley~J Maddox, Jayson~P Salkey, J{\'u}lio Albinati, and
  Andrew~Gordon Wilson.
\newblock {Function-space distributions over kernels}.
\newblock \emph{Advances in Neural Information Processing Systems}, 32, 2019.

\bibitem[Maro{\~n}as et~al.(2021)Maro{\~n}as, Hamelijnck, Knoblauch, and
  Damoulas]{maronas2021transforming}
Juan Maro{\~n}as, Oliver Hamelijnck, Jeremias Knoblauch, and Theodoros
  Damoulas.
\newblock {Transforming Gaussian processes with normalizing flows}.
\newblock In \emph{International Conference on Artificial Intelligence and
  Statistics}, pages 1081--1089. PMLR, 2021.

\bibitem[Wilson and Ghahramani(2010)]{wilson2010copula}
Andrew~Gordon Wilson and Zoubin Ghahramani.
\newblock {Copula processes}.
\newblock In \emph{Proceedings of the 23rd International Conference on Neural
  Information Processing Systems-Volume 2}, pages 2460--2468, 2010.

\bibitem[Neal(1995)]{neal1995bayesian}
Radford~M Neal.
\newblock \emph{{Bayesian learning for neural networks}}.
\newblock PhD thesis, University of Toronto, 1995.

\bibitem[Williams(1996)]{williams1996computing}
Christopher~KI Williams.
\newblock {Computing with infinite networks}.
\newblock In \emph{{Proceedings of the 9th International Conference on Neural
  Information Processing Systems}}, pages 295--301, 1996.

\bibitem[Cho and Saul(2009)]{cho2009kernel}
Youngmin Cho and Lawrence~K Saul.
\newblock {Kernel methods for deep learning}.
\newblock In \emph{{Proceedings of the 22nd International Conference on Neural
  Information Processing Systems}}, pages 342--350, 2009.

\bibitem[Hazan and Jaakkola(2015)]{hazan2015steps}
Tamir Hazan and Tommi Jaakkola.
\newblock {Steps toward deep kernel methods from infinite neural networks}.
\newblock \emph{arXiv preprint arXiv:1508.05133}, 2015.

\bibitem[Lee et~al.(2018)Lee, Bahri, Novak, Schoenholz, Pennington, and
  Sohl-Dickstein]{lee2018deep}
Jaehoon Lee, Yasaman Bahri, Roman Novak, Samuel~S Schoenholz, Jeffrey
  Pennington, and Jascha Sohl-Dickstein.
\newblock {Deep Neural Networks as Gaussian Processes}.
\newblock In \emph{{International Conference on Learning Representations}},
  2018.

\bibitem[Matthews et~al.(2018)Matthews, Hron, Rowland, Turner, and
  Ghahramani]{matthews2018gaussian}
Alexander G de~G Matthews, Jiri Hron, Mark Rowland, Richard~E Turner, and
  Zoubin Ghahramani.
\newblock {Gaussian Process Behaviour in Wide Deep Neural Networks}.
\newblock In \emph{{International Conference on Learning Representations}},
  2018.

\bibitem[Garriga{-}Alonso et~al.(2019)Garriga{-}Alonso, Rasmussen, and
  Aitchison]{garriga2018deep}
Adri{\`{a}} Garriga{-}Alonso, Carl~Edward Rasmussen, and Laurence Aitchison.
\newblock {Deep Convolutional Networks as shallow Gaussian Processes}.
\newblock In \emph{7th International Conference on Learning Representations},
  2019.

\bibitem[Novak et~al.(2019{\natexlab{a}})Novak, Xiao, Bahri, Lee, Yang,
  Abolafia, Pennington, and Sohl-dickstein]{novak2019bayesian}
Roman Novak, Lechao Xiao, Yasaman Bahri, Jaehoon Lee, Greg Yang, Daniel~A.
  Abolafia, Jeffrey Pennington, and Jascha Sohl-dickstein.
\newblock {Bayesian Deep Convolutional Networks with Many Channels are Gaussian
  Processes}.
\newblock In \emph{International Conference on Learning Representations},
  2019{\natexlab{a}}.

\bibitem[Garriga-Alonso and van~der Wilk(2021)]{garriga2021correlated}
Adri{\`a} Garriga-Alonso and Mark van~der Wilk.
\newblock {Correlated Weights in Infinite Limits of Deep Convolutional Neural
  Networks}.
\newblock \emph{arXiv preprint arXiv:2101.04097}, 2021.

\bibitem[Hron et~al.(2020{\natexlab{a}})Hron, Bahri, Sohl-Dickstein, and
  Novak]{hron2020infinite}
Jiri Hron, Yasaman Bahri, Jascha Sohl-Dickstein, and Roman Novak.
\newblock {Infinite attention: NNGP and NTK for deep attention networks}.
\newblock In \emph{{International Conference on Machine Learning}}, pages
  4376--4386. PMLR, 2020{\natexlab{a}}.

\bibitem[Tsuchida et~al.(2019)Tsuchida, Roosta, and
  Gallagher]{tsuchida2019richer}
Russell Tsuchida, Fred Roosta, and Marcus Gallagher.
\newblock {Richer priors for infinitely wide multi-layer perceptrons}.
\newblock \emph{arXiv preprint arXiv:1911.12927}, 2019.

\bibitem[Peluchetti et~al.(2020)Peluchetti, Favaro, and
  Fortini]{peluchetti2020stable}
Stefano Peluchetti, Stefano Favaro, and Sandra Fortini.
\newblock {Stable behaviour of infinitely wide deep neural networks}.
\newblock In \emph{International Conference on Artificial Intelligence and
  Statistics}, pages 1137--1146. PMLR, 2020.

\bibitem[Hron et~al.(2020{\natexlab{b}})Hron, Bahri, Novak, Pennington, and
  Sohl-Dickstein]{hron2020exact}
Jiri Hron, Yasaman Bahri, Roman Novak, Jeffrey Pennington, and Jascha
  Sohl-Dickstein.
\newblock {Exact posterior distributions of wide Bayesian neural networks}.
\newblock \emph{arXiv preprint arXiv:2006.10541}, 2020{\natexlab{b}}.

\bibitem[Yang(2019{\natexlab{a}})]{yang2019scaling}
Greg Yang.
\newblock {Scaling limits of wide neural networks with weight sharing: Gaussian
  process behavior, gradient independence, and neural tangent kernel
  derivation}.
\newblock \emph{arXiv preprint arXiv:1902.04760}, 2019{\natexlab{a}}.

\bibitem[Yang(2019{\natexlab{b}})]{yang2019tensori}
Greg Yang.
\newblock {Tensor programs i: Wide feedforward or recurrent neural networks of
  any architecture are gaussian processes}.
\newblock \emph{arXiv preprint arXiv:1910.12478}, 2019{\natexlab{b}}.

\bibitem[Yang(2020{\natexlab{a}})]{yang2020tensoriii}
Greg Yang.
\newblock {Tensor programs iii: Neural matrix laws}.
\newblock \emph{arXiv preprint arXiv:2009.10685}, 2020{\natexlab{a}}.

\bibitem[Yang and Hu(2020)]{yang2020feature}
Greg Yang and Edward~J Hu.
\newblock {Feature Learning in Infinite-Width Neural Networks}.
\newblock \emph{arXiv preprint arXiv:2011.14522}, 2020.

\bibitem[Jacot et~al.(2018)Jacot, Gabriel, and Hongler]{jacot2018neural}
Arthur Jacot, Franck Gabriel, and Cl{\'e}ment Hongler.
\newblock {Neural tangent kernel: convergence and generalization in neural
  networks}.
\newblock In \emph{{Proceedings of the 32nd International Conference on Neural
  Information Processing Systems}}, pages 8580--8589, 2018.

\bibitem[Lee et~al.(2020)Lee, Xiao, Schoenholz, Bahri, Novak, Sohl-Dickstein,
  and Pennington]{lee2020wide}
Jaehoon Lee, Lechao Xiao, Samuel~S Schoenholz, Yasaman Bahri, Roman Novak,
  Jascha Sohl-Dickstein, and Jeffrey Pennington.
\newblock {Wide neural networks of any depth evolve as linear models under
  gradient descent}.
\newblock \emph{Journal of Statistical Mechanics: Theory and Experiment},
  2020\penalty0 (12):\penalty0 124002, 2020.

\bibitem[Hanin and Nica(2019)]{hanin2019finite}
Boris Hanin and Mihai Nica.
\newblock {Finite Depth and Width Corrections to the Neural Tangent Kernel}.
\newblock In \emph{{International Conference on Learning Representations}},
  2019.

\bibitem[Khan et~al.(2019)Khan, Immer, Abedi, and Korzepa]{khan2019approximate}
Mohammad~Emtiyaz Khan, Alexander Immer, Ehsan Abedi, and Maciej~Jan Korzepa.
\newblock {Approximate Inference Turns Deep Networks into Gaussian Processes}.
\newblock In \emph{{33rd Conference on Neural Information Processing Systems}},
  page 1751. Neural Information Processing Systems Foundation, 2019.

\bibitem[Immer et~al.(2021{\natexlab{a}})Immer, Korzepa, and
  Bauer]{immer2021improving}
Alexander Immer, Maciej Korzepa, and Matthias Bauer.
\newblock {Improving predictions of Bayesian neural nets via local
  linearization}.
\newblock In \emph{International Conference on Artificial Intelligence and
  Statistics}, pages 703--711. PMLR, 2021{\natexlab{a}}.

\bibitem[Maddox et~al.(2021)Maddox, Tang, Moreno, Wilson, and
  Damianou]{maddox2021fast}
Wesley Maddox, Shuai Tang, Pablo Moreno, Andrew~Gordon Wilson, and Andreas
  Damianou.
\newblock {Fast Adaptation with Linearized Neural Networks}.
\newblock In \emph{{International Conference on Artificial Intelligence and
  Statistics}}, pages 2737--2745. PMLR, 2021.

\bibitem[Arora et~al.(2019)Arora, Du, Li, Salakhutdinov, Wang, and
  Yu]{arora2019harnessing}
Sanjeev Arora, Simon~S Du, Zhiyuan Li, Ruslan Salakhutdinov, Ruosong Wang, and
  Dingli Yu.
\newblock {Harnessing the Power of Infinitely Wide Deep Nets on Small-data
  Tasks}.
\newblock In \emph{{International Conference on Learning Representations}},
  2019.

\bibitem[Yang(2020{\natexlab{b}})]{yang2020tensorii}
Greg Yang.
\newblock {Tensor programs ii: Neural tangent kernel for any architecture}.
\newblock \emph{arXiv preprint arXiv:2006.14548}, 2020{\natexlab{b}}.

\bibitem[Novak et~al.(2019{\natexlab{b}})Novak, Xiao, Hron, Lee, Alemi,
  Sohl-Dickstein, and Schoenholz]{novak2019neural}
Roman Novak, Lechao Xiao, Jiri Hron, Jaehoon Lee, Alexander~A Alemi, Jascha
  Sohl-Dickstein, and Samuel~S Schoenholz.
\newblock {Neural Tangents: Fast and Easy Infinite Neural Networks in Python}.
\newblock In \emph{{International Conference on Learning Representations}},
  2019{\natexlab{b}}.

\bibitem[Chizat et~al.(2019)Chizat, Oyallon, and Bach]{chizat2019lazy}
L{\'e}na{\"\i}c Chizat, Edouard Oyallon, and Francis Bach.
\newblock {On Lazy Training in Differentiable Programming}.
\newblock \emph{Advances in Neural Information Processing Systems},
  32:\penalty0 2937--2947, 2019.

\bibitem[Kingma and Welling(2014)]{kingma2013auto}
Diederik~P Kingma and Max Welling.
\newblock {Auto-encoding variational bayes}.
\newblock \emph{International Conference on Learning Representations}, 2014.

\bibitem[Rezende et~al.(2014)Rezende, Mohamed, and
  Wierstra]{rezende2014stochastic}
Danilo~Jimenez Rezende, Shakir Mohamed, and Daan Wierstra.
\newblock {Stochastic backpropagation and approximate inference in deep
  generative models}.
\newblock In \emph{{International conference on machine learning}}, pages
  1278--1286. PMLR, 2014.

\bibitem[Davidson et~al.(2018)Davidson, Falorsi, De~Cao, Kipf, and
  Tomczak]{davidson2018hyperspherical}
Tim~R Davidson, Luca Falorsi, Nicola De~Cao, Thomas Kipf, and Jakub~M Tomczak.
\newblock {Hyperspherical variational auto-encoders}.
\newblock In \emph{{34th Conference on Uncertainty in Artificial Intelligence
  2018, UAI 2018}}, pages 856--865. Association For Uncertainty in Artificial
  Intelligence (AUAI), 2018.

\bibitem[Blum et~al.(2020)Blum, Hopcroft, and Kannan]{blum2020foundations}
Avrim Blum, John Hopcroft, and Ravindran Kannan.
\newblock \emph{{Foundations of data science}}.
\newblock Cambridge University Press, 2020.

\bibitem[Davidson et~al.(2019)Davidson, Tomczak, and
  Gavves]{davidson2019increasing}
Tim~R Davidson, Jakub~M Tomczak, and Efstratios Gavves.
\newblock {Increasing Expressivity of a Hyperspherical VAE}.
\newblock \emph{arXiv preprint arXiv:1910.02912}, 2019.

\bibitem[{De Cao} and Aziz(2020)]{de2020power}
Nicola {De Cao} and Wilker Aziz.
\newblock {The power spherical distribution}.
\newblock \emph{arXiv preprint arXiv:2006.04437}, 2020.

\bibitem[Dilokthanakul et~al.(2016)Dilokthanakul, Mediano, Garnelo, Lee,
  Salimbeni, Arulkumaran, and Shanahan]{dilokthanakul2016deep}
Nat Dilokthanakul, Pedro~AM Mediano, Marta Garnelo, Matthew~CH Lee, Hugh
  Salimbeni, Kai Arulkumaran, and Murray Shanahan.
\newblock {Deep unsupervised clustering with gaussian mixture variational
  autoencoders}.
\newblock \emph{arXiv preprint arXiv:1611.02648}, 2016.

\bibitem[Jiang et~al.(2017)Jiang, Zheng, Tan, Tang, and
  Zhou]{jiang2017variational}
Zhuxi Jiang, Yin Zheng, Huachun Tan, Bangsheng Tang, and Hanning Zhou.
\newblock {Variational deep embedding: an unsupervised and generative approach
  to clustering}.
\newblock In \emph{{Proceedings of the 26th International Joint Conference on
  Artificial Intelligence}}, pages 1965--1972, 2017.

\bibitem[Kopf et~al.(2021)Kopf, Fortuin, Somnath, and
  Claassen]{kopf2019mixture}
Andreas Kopf, Vincent Fortuin, Vignesh~Ram Somnath, and Manfred Claassen.
\newblock {Mixture-of-Experts Variational Autoencoder for clustering and
  generating from similarity-based representations}.
\newblock \emph{PLoS Computational Biology}, 2021.

\bibitem[Nalisnick and Smyth(2016)]{nalisnick2016stick}
Eric Nalisnick and Padhraic Smyth.
\newblock {Stick-breaking variational autoencoders}.
\newblock \emph{arXiv preprint arXiv:1605.06197}, 2016.

\bibitem[Casale et~al.(2018)Casale, Dalca, Saglietti, Listgarten, and
  Fusi]{casale2018gaussian}
Francesco~Paolo Casale, Adrian Dalca, Luca Saglietti, Jennifer Listgarten, and
  Nicolo Fusi.
\newblock {Gaussian Process Prior Variational Autoencoders}.
\newblock In \emph{Advances in Neural Information Processing Systems}, pages
  10369--10380, 2018.

\bibitem[Fortuin et~al.(2020)Fortuin, Baranchuk, R{\"a}tsch, and
  Mandt]{fortuin2020gp}
Vincent Fortuin, Dmitry Baranchuk, Gunnar R{\"a}tsch, and Stephan Mandt.
\newblock {GP-VAE: Deep Probabilistic Time Series Imputation}.
\newblock In \emph{{International Conference on Artificial Intelligence and
  Statistics}}, pages 1651--1661. PMLR, 2020.

\bibitem[Pearce(2020)]{pearce2020gaussian}
Michael Pearce.
\newblock {The gaussian process prior vae for interpretable latent dynamics
  from pixels}.
\newblock In \emph{{Symposium on Advances in Approximate Bayesian Inference}},
  pages 1--12. PMLR, 2020.

\bibitem[Bhagat et~al.(2020)Bhagat, Uppal, Yin, and
  Lim]{bhagat2020disentangling}
Sarthak Bhagat, Shagun Uppal, Zhuyun Yin, and Nengli Lim.
\newblock {Disentangling Multiple Features in Video Sequences using Gaussian
  Processes in Variational Autoencoders}.
\newblock In \emph{{European Conference on Computer Vision}}, pages 102--117.
  Springer, 2020.

\bibitem[Bing et~al.(2021)Bing, Fortuin, and
  R{\"a}tsch]{bing2021disentanglement}
Simon Bing, Vincent Fortuin, and Gunnar R{\"a}tsch.
\newblock {On Disentanglement in Gaussian Process Variational Autoencoders}.
\newblock \emph{arXiv preprint arXiv:2102.05507}, 2021.

\bibitem[Jazbec et~al.(2021)Jazbec, Ashman, Fortuin, Pearce, Mandt, and
  R{\"a}tsch]{jazbec2020scalable}
Metod Jazbec, Matthew Ashman, Vincent Fortuin, Michael Pearce, Stephan Mandt,
  and Gunnar R{\"a}tsch.
\newblock {Scalable Gaussian Process Variational Autoencoders}.
\newblock In \emph{{International Conference on Artificial Intelligence and
  Statistics}}, 2021.

\bibitem[Ashman et~al.(2020)Ashman, So, Tebbutt, Fortuin, Pearce, and
  Turner]{ashman2020sparse}
Matthew Ashman, Jonathan So, William Tebbutt, Vincent Fortuin, Michael Pearce,
  and Richard~E Turner.
\newblock {Sparse Gaussian Process Variational Autoencoders}.
\newblock \emph{arXiv preprint arXiv:2010.10177}, 2020.

\bibitem[Jazbec et~al.(2020)Jazbec, Pearce, and Fortuin]{jazbec2020factorized}
Metod Jazbec, Michael Pearce, and Vincent Fortuin.
\newblock {Factorized Gaussian Process Variational Autoencoders}.
\newblock \emph{arXiv preprint arXiv:2011.07255}, 2020.

\bibitem[Ramchandran et~al.(2020)Ramchandran, Tikhonov, Koskinen, and
  L{\"a}hdesm{\"a}ki]{ramchandran2020longitudinal}
Siddharth Ramchandran, Gleb Tikhonov, Miika Koskinen, and Harri
  L{\"a}hdesm{\"a}ki.
\newblock {Longitudinal Variational Autoencoder}.
\newblock \emph{arXiv preprint arXiv:2006.09763}, 2020.

\bibitem[Campbell and Li{\`o}(2020)]{campbell2020tvgp}
Alex Campbell and Pietro Li{\`o}.
\newblock {tvGP-VAE: Tensor-variate Gaussian process prior variational
  autoencoder}.
\newblock \emph{arXiv preprint arXiv:2006.04788}, 2020.

\bibitem[S{\o}nderby et~al.(2016)S{\o}nderby, Raiko, Maal{\o}e, S{\o}nderby,
  and Winther]{sonderby2016ladder}
Casper~Kaae S{\o}nderby, Tapani Raiko, Lars Maal{\o}e, S{\o}ren~Kaae
  S{\o}nderby, and Ole Winther.
\newblock {Ladder variational autoencoders}.
\newblock In \emph{{Proceedings of the 30th International Conference on Neural
  Information Processing Systems}}, pages 3745--3753, 2016.

\bibitem[Zhao et~al.(2017)Zhao, Song, and Ermon]{zhao2017learning}
Shengjia Zhao, Jiaming Song, and Stefano Ermon.
\newblock {Learning hierarchical features from deep generative models}.
\newblock In \emph{{International Conference on Machine Learning}}, pages
  4091--4099. PMLR, 2017.

\bibitem[Vahdat and Kautz(2020)]{vahdat2020nvae}
Arash Vahdat and Jan Kautz.
\newblock {NVAE: A deep hierarchical variational autoencoder}.
\newblock \emph{arXiv preprint arXiv:2007.03898}, 2020.

\bibitem[{van den Oord} et~al.(2017){van den Oord}, Vinyals, and
  Kavukcuoglu]{van2017neural}
Aaron {van den Oord}, Oriol Vinyals, and Koray Kavukcuoglu.
\newblock {Neural discrete representation learning}.
\newblock In \emph{{Proceedings of the 31st International Conference on Neural
  Information Processing Systems}}, pages 6309--6318, 2017.

\bibitem[Razavi et~al.(2019)Razavi, Oord, and Vinyals]{razavi2019generating}
Ali Razavi, Aaron van~den Oord, and Oriol Vinyals.
\newblock {Generating diverse high-fidelity images with VQ-VAE-2}.
\newblock \emph{arXiv preprint arXiv:1906.00446}, 2019.

\bibitem[Kohonen(1990)]{Kohonen1998}
Teuvo Kohonen.
\newblock {The self-organizing map}.
\newblock \emph{Proceedings of the IEEE}, 78\penalty0 (9):\penalty0 1464--1480,
  1990.

\bibitem[Fortuin et~al.(2019{\natexlab{b}})Fortuin, H{\"u}ser, Locatello,
  Strathmann, and R{\"a}tsch]{fortuin2019som}
Vincent Fortuin, Matthias H{\"u}ser, Francesco Locatello, Heiko Strathmann, and
  Gunnar R{\"a}tsch.
\newblock {SOM-VAE: Interpretable Discrete Representation Learning on Time
  Series}.
\newblock In \emph{{International Conference on Learning Representations}},
  2019{\natexlab{b}}.

\bibitem[Forest et~al.(2019)Forest, Lebbah, Azzag, and
  Lacaille]{forest2019deep}
Florent Forest, Mustapha Lebbah, Hanane Azzag, and J{\'e}r{\^o}me Lacaille.
\newblock {Deep architectures for joint clustering and visualization with
  self-organizing maps}.
\newblock In \emph{{Pacific-Asia Conference on Knowledge Discovery and Data
  Mining}}, pages 105--116. Springer, 2019.

\bibitem[Manduchi et~al.(2019)Manduchi, H{\"u}ser, Vogt, R{\"a}tsch, and
  Fortuin]{manduchi2019dpsom}
Laura Manduchi, Matthias H{\"u}ser, Julia Vogt, Gunnar R{\"a}tsch, and Vincent
  Fortuin.
\newblock {DPSOM: Deep probabilistic clustering with self-organizing maps}.
\newblock \emph{arXiv preprint arXiv:1910.01590}, 2019.

\bibitem[Moor et~al.(2020)Moor, Horn, Rieck, and
  Borgwardt]{moor2020topological}
Michael Moor, Max Horn, Bastian Rieck, and Karsten Borgwardt.
\newblock {Topological autoencoders}.
\newblock In \emph{International conference on machine learning}, pages
  7045--7054. PMLR, 2020.

\bibitem[Gabrielsson et~al.(2020)Gabrielsson, Nelson, Dwaraknath, and
  Skraba]{gabrielsson2020topology}
Rickard~Br{\"u}el Gabrielsson, Bradley~J Nelson, Anjan Dwaraknath, and Primoz
  Skraba.
\newblock {A topology layer for machine learning}.
\newblock In \emph{International Conference on Artificial Intelligence and
  Statistics}, pages 1553--1563. PMLR, 2020.

\bibitem[Ma et~al.(2019{\natexlab{a}})Ma, Tschiatschek, Palla,
  Hernandez-Lobato, Nowozin, and Zhang]{ma2019eddi}
Chao Ma, Sebastian Tschiatschek, Konstantina Palla, Jose~Miguel
  Hernandez-Lobato, Sebastian Nowozin, and Cheng Zhang.
\newblock {EDDI: Efficient Dynamic Discovery of High-Value Information with
  Partial VAE}.
\newblock In \emph{{International Conference on Machine Learning}}, pages
  4234--4243. PMLR, 2019{\natexlab{a}}.

\bibitem[Garnelo et~al.(2018{\natexlab{a}})Garnelo, Rosenbaum, Maddison,
  Ramalho, Saxton, Shanahan, Teh, Rezende, and Eslami]{garnelo2018conditional}
Marta Garnelo, Dan Rosenbaum, Christopher Maddison, Tiago Ramalho, David
  Saxton, Murray Shanahan, Yee~Whye Teh, Danilo Rezende, and SM~Ali Eslami.
\newblock {Conditional Neural Processes}.
\newblock In \emph{International Conference on Machine Learning}, pages
  1690--1699, 2018{\natexlab{a}}.

\bibitem[Garnelo et~al.(2018{\natexlab{b}})Garnelo, Schwarz, Rosenbaum, Viola,
  Rezende, Eslami, and Teh]{garnelo2018neural}
Marta Garnelo, Jonathan Schwarz, Dan Rosenbaum, Fabio Viola, Danilo~J Rezende,
  SM~Eslami, and Yee~Whye Teh.
\newblock {Neural processes}.
\newblock \emph{arXiv preprint arXiv:1807.01622}, 2018{\natexlab{b}}.

\bibitem[Sohn et~al.(2015)Sohn, Yan, and Lee]{sohn2015learning}
Kihyuk Sohn, Xinchen Yan, and Honglak Lee.
\newblock {Learning structured output representation using deep conditional
  generative models}.
\newblock In \emph{{Proceedings of the 28th International Conference on Neural
  Information Processing Systems-Volume 2}}, pages 3483--3491, 2015.

\bibitem[Rivasplata et~al.(2020)Rivasplata, Kuzborskij, Szepesv{\'a}ri, and
  Shawe-Taylor]{rivasplata2020pac}
Omar Rivasplata, Ilja Kuzborskij, Csaba Szepesv{\'a}ri, and John Shawe-Taylor.
\newblock {PAC-Bayes analysis beyond the usual bounds}.
\newblock \emph{arXiv preprint arXiv:2006.13057}, 2020.

\bibitem[Dziugaite et~al.(2021)Dziugaite, Hsu, Gharbieh, Arpino, and
  Roy]{dziugaite2021role}
Gintare~Karolina Dziugaite, Kyle Hsu, Waseem Gharbieh, Gabriel Arpino, and
  Daniel Roy.
\newblock {On the role of data in PAC-Bayes}.
\newblock In \emph{International Conference on Artificial Intelligence and
  Statistics}, pages 604--612. PMLR, 2021.

\bibitem[Rudner et~al.(2018)Rudner, Fortuin, Teh, and
  Gal]{rudner2018connection}
Tim~GJ Rudner, Vincent Fortuin, Yee~Whye Teh, and Yarin Gal.
\newblock {On the connection between neural processes and Gaussian processes
  with deep kernels}.
\newblock In \emph{{Workshop on Bayesian Deep Learning, NeurIPS}}, 2018.

\bibitem[Kim et~al.(2018)Kim, Mnih, Schwarz, Garnelo, Eslami, Rosenbaum,
  Vinyals, and Teh]{kim2018attentive}
Hyunjik Kim, Andriy Mnih, Jonathan Schwarz, Marta Garnelo, Ali Eslami, Dan
  Rosenbaum, Oriol Vinyals, and Yee~Whye Teh.
\newblock {Attentive Neural Processes}.
\newblock In \emph{{International Conference on Learning Representations}},
  2018.

\bibitem[P{\'e}rez et~al.(2021)P{\'e}rez, Barcel{\'o}, and
  Marinkovic]{perez2021attention}
Jorge P{\'e}rez, Pablo Barcel{\'o}, and Javier Marinkovic.
\newblock {Attention is Turing-Complete}.
\newblock \emph{Journal of Machine Learning Research}, 22\penalty0
  (75):\penalty0 1--35, 2021.

\bibitem[Gordon et~al.(2019)Gordon, Bruinsma, Foong, Requeima, Dubois, and
  Turner]{gordon2019convolutional}
Jonathan Gordon, Wessel~P Bruinsma, Andrew~YK Foong, James Requeima, Yann
  Dubois, and Richard~E Turner.
\newblock {Convolutional Conditional Neural Processes}.
\newblock In \emph{{International Conference on Learning Representations}},
  2019.

\bibitem[Foong et~al.(2020)Foong, Bruinsma, Gordon, Dubois, Requeima, and
  Turner]{foong2020meta}
Andrew Foong, Wessel Bruinsma, Jonathan Gordon, Yann Dubois, James Requeima,
  and Richard Turner.
\newblock {Meta-Learning Stationary Stochastic Process Prediction with
  Convolutional Neural Processes}.
\newblock \emph{Advances in Neural Information Processing Systems}, 33, 2020.

\bibitem[Louizos et~al.(2019)Louizos, Shi, Schutte, and
  Welling]{louizos2019functional}
Christos Louizos, Xiahan Shi, Klamer Schutte, and Max Welling.
\newblock {The functional neural process}.
\newblock \emph{arXiv preprint arXiv:1906.08324}, 2019.

\bibitem[Bruinsma et~al.(2021)Bruinsma, Requeima, Foong, Gordon, and
  Turner]{bruinsma2021gaussian}
Wessel~P Bruinsma, James Requeima, Andrew~YK Foong, Jonathan Gordon, and
  Richard~E Turner.
\newblock {The Gaussian Neural Process}.
\newblock \emph{arXiv preprint arXiv:2101.03606}, 2021.

\bibitem[MacKay(1992)]{mackay1992practical}
David~J.C. MacKay.
\newblock {A practical Bayesian framework for backpropagation networks}.
\newblock \emph{Neural computation}, 4\penalty0 (3):\penalty0 448--472, 1992.

\bibitem[Jospin et~al.(2020)Jospin, Buntine, Boussaid, Laga, and
  Bennamoun]{jospin2020hands}
Laurent~Valentin Jospin, Wray Buntine, Farid Boussaid, Hamid Laga, and Mohammed
  Bennamoun.
\newblock {Hands-on Bayesian Neural Networks--a Tutorial for Deep Learning
  Users}.
\newblock \emph{arXiv preprint arXiv:2007.06823}, 2020.

\bibitem[Nalisnick(2018)]{nalisnick2018priors}
Eric~T Nalisnick.
\newblock \emph{{On priors for Bayesian neural networks}}.
\newblock PhD thesis, UC Irvine, 2018.

\bibitem[Wilson and Izmailov(2020)]{wilson2020bayesian}
Andrew~Gordon Wilson and Pavel Izmailov.
\newblock {Bayesian deep learning and a probabilistic perspective of
  generalization}.
\newblock \emph{arXiv preprint arXiv:2002.08791}, 2020.

\bibitem[Silvestro and Andermann(2020)]{silvestro2020prior}
Daniele Silvestro and Tobias Andermann.
\newblock {Prior choice affects ability of Bayesian neural networks to identify
  unknowns}.
\newblock \emph{arXiv preprint arXiv:2005.04987}, 2020.

\bibitem[Hern{\'a}ndez-Lobato and Adams(2015)]{hernandez2015probabilistic}
Jos{\'e}~Miguel Hern{\'a}ndez-Lobato and Ryan Adams.
\newblock {Probabilistic backpropagation for scalable learning of bayesian
  neural networks}.
\newblock In \emph{{International Conference on Machine Learning}}, pages
  1861--1869. PMLR, 2015.

\bibitem[Louizos and Welling(2017)]{louizos2017multiplicative}
Christos Louizos and Max Welling.
\newblock {Multiplicative normalizing flows for variational bayesian neural
  networks}.
\newblock In \emph{{International Conference on Machine Learning}}, pages
  2218--2227. PMLR, 2017.

\bibitem[Zhang et~al.(2019)Zhang, Li, Zhang, Chen, and
  Wilson]{zhang2019cyclical}
Ruqi Zhang, Chunyuan Li, Jianyi Zhang, Changyou Chen, and Andrew~Gordon Wilson.
\newblock {Cyclical Stochastic Gradient MCMC for Bayesian Deep Learning}.
\newblock In \emph{{International Conference on Learning Representations}},
  2019.

\bibitem[Dusenberry et~al.(2020)Dusenberry, Jerfel, Wen, Ma, Snoek, Heller,
  Lakshminarayanan, and Tran]{dusenberry2020efficient}
Michael Dusenberry, Ghassen Jerfel, Yeming Wen, Yian Ma, Jasper Snoek,
  Katherine Heller, Balaji Lakshminarayanan, and Dustin Tran.
\newblock {Efficient and scalable bayesian neural nets with rank-1 factors}.
\newblock In \emph{{International conference on machine learning}}, pages
  2782--2792. PMLR, 2020.

\bibitem[Vladimirova et~al.(2019)Vladimirova, Verbeek, Mesejo, and
  Arbel]{vladimirova2019understanding}
Mariia Vladimirova, Jakob Verbeek, Pablo Mesejo, and Julyan Arbel.
\newblock {Understanding priors in Bayesian neural networks at the unit level}.
\newblock In \emph{International Conference on Machine Learning}, pages
  6458--6467. PMLR, 2019.

\bibitem[MacKay(1998)]{mackay1998introduction}
David~JC MacKay.
\newblock {Introduction to Gaussian processes}.
\newblock \emph{NATO ASI series F computer and systems sciences}, 168:\penalty0
  133--166, 1998.

\bibitem[Louizos and Welling(2016)]{louizos2016structured}
Christos Louizos and Max Welling.
\newblock {Structured and efficient variational deep learning with matrix
  gaussian posteriors}.
\newblock In \emph{{International Conference on Machine Learning}}, pages
  1708--1716. PMLR, 2016.

\bibitem[Graves(2011)]{graves2011practical}
Alex Graves.
\newblock {Practical variational inference for neural networks}.
\newblock In \emph{{Advances in neural information processing systems}}, pages
  2348--2356. Citeseer, 2011.

\bibitem[Wu et~al.(2018)Wu, Nowozin, Meeds, Turner, Hern{\'a}ndez-Lobato, and
  Gaunt]{wu2018deterministic}
Anqi Wu, Sebastian Nowozin, Edward Meeds, Richard~E Turner, Jos{\'e}~Miguel
  Hern{\'a}ndez-Lobato, and Alexander~L Gaunt.
\newblock {Deterministic Variational Inference for Robust Bayesian Neural
  Networks}.
\newblock In \emph{{International Conference on Learning Representations}},
  2018.

\bibitem[Shah et~al.(2014)Shah, Wilson, and Ghahramani]{shah2014student}
Amar Shah, Andrew Wilson, and Zoubin Ghahramani.
\newblock {Student-t processes as alternatives to Gaussian processes}.
\newblock In \emph{{Artificial intelligence and statistics}}, pages 877--885.
  PMLR, 2014.

\bibitem[Williams(1995)]{williams1995bayesian}
Peter~M Williams.
\newblock {Bayesian regularization and pruning using a Laplace prior}.
\newblock \emph{Neural computation}, 7\penalty0 (1):\penalty0 117--143, 1995.

\bibitem[Neklyudov et~al.(2018)Neklyudov, Molchanov, Ashukha, and
  Vetrov]{neklyudov2018variance}
Kirill Neklyudov, Dmitry Molchanov, Arsenii Ashukha, and Dmitry Vetrov.
\newblock {Variance Networks: When Expectation Does Not Meet Your
  Expectations}.
\newblock In \emph{{International Conference on Learning Representations}},
  2018.

\bibitem[Carvalho et~al.(2009)Carvalho, Polson, and
  Scott]{carvalho2009handling}
Carlos~M Carvalho, Nicholas~G Polson, and James~G Scott.
\newblock {Handling sparsity via the horseshoe}.
\newblock In \emph{{Artificial Intelligence and Statistics}}, pages 73--80.
  PMLR, 2009.

\bibitem[Ghosh et~al.(2018)Ghosh, Yao, and Doshi-Velez]{ghosh2018structured}
Soumya Ghosh, Jiayu Yao, and Finale Doshi-Velez.
\newblock {Structured variational learning of Bayesian neural networks with
  horseshoe priors}.
\newblock In \emph{{International Conference on Machine Learning}}, pages
  1744--1753. PMLR, 2018.

\bibitem[Overweg et~al.(2019)Overweg, Popkes, Ercole, Li, Hern{\'a}ndez-Lobato,
  Zaykov, and Zhang]{overweg2019interpretable}
Hiske Overweg, Anna-Lena Popkes, Ari Ercole, Yingzhen Li, Jos{\'e}~Miguel
  Hern{\'a}ndez-Lobato, Yordan Zaykov, and Cheng Zhang.
\newblock {Interpretable Outcome Prediction with Sparse Bayesian Neural
  Networks in Intensive Care}.
\newblock \emph{arXiv preprint arXiv:1905.02599}, 2019.

\bibitem[Louizos et~al.(2017)Louizos, Ullrich, and
  Welling]{louizos2017bayesian}
Christos Louizos, Karen Ullrich, and Max Welling.
\newblock {Bayesian compression for deep learning}.
\newblock In \emph{{Proceedings of the 31st International Conference on Neural
  Information Processing Systems}}, pages 3290--3300, 2017.

\bibitem[Cui et~al.(2020)Cui, Havulinna, Marttinen, and
  Kaski]{cui2020informative}
Tianyu Cui, A.~Havulinna, P.~Marttinen, and S.~Kaski.
\newblock {Informative Gaussian Scale Mixture Priors for Bayesian Neural
  Networks}.
\newblock \emph{arXiv preprint arXiv:2002.10243}, 2020.

\bibitem[Srivastava et~al.(2014)Srivastava, Hinton, Krizhevsky, Sutskever, and
  Salakhutdinov]{srivastava2014dropout}
Nitish Srivastava, Geoffrey Hinton, Alex Krizhevsky, Ilya Sutskever, and Ruslan
  Salakhutdinov.
\newblock {Dropout: a simple way to prevent neural networks from overfitting}.
\newblock \emph{The journal of machine learning research}, 15\penalty0
  (1):\penalty0 1929--1958, 2014.

\bibitem[Nalisnick et~al.(2019)Nalisnick, Hern{\'a}ndez-Lobato, and
  Smyth]{nalisnick2019dropout}
Eric Nalisnick, Jos{\'e}~Miguel Hern{\'a}ndez-Lobato, and Padhraic Smyth.
\newblock {Dropout as a structured shrinkage prior}.
\newblock In \emph{International Conference on Machine Learning}, pages
  4712--4722. PMLR, 2019.

\bibitem[Hubin and Storvik(2019)]{hubin2019combining}
Aliaksandr Hubin and Geir Storvik.
\newblock {Combining model and parameter uncertainty in Bayesian neural
  networks}.
\newblock \emph{arXiv preprint arXiv:1903.07594}, 2019.

\bibitem[Kessler et~al.(2019)Kessler, Nguyen, Zohren, and
  Roberts]{kessler2019hierarchical}
Samuel Kessler, Vu~Nguyen, Stefan Zohren, and Stephen Roberts.
\newblock {Hierarchical Indian Buffet Neural Networks for Bayesian Continual
  Learning}.
\newblock \emph{arXiv preprint arXiv:1912.02290}, 2019.

\bibitem[Oh et~al.(2019)Oh, Adamczewski, and Park]{oh2019radial}
Changyong Oh, Kamil Adamczewski, and Mijung Park.
\newblock {Radial and directional posteriors for bayesian neural networks}.
\newblock \emph{arXiv preprint arXiv:1902.02603}, 2019.

\bibitem[Farquhar et~al.(2020)Farquhar, Osborne, and Gal]{farquhar2020radial}
Sebastian Farquhar, Michael~A Osborne, and Yarin Gal.
\newblock {Radial bayesian neural networks: Beyond discrete support in
  large-scale bayesian deep learning}.
\newblock In \emph{{International Conference on Artificial Intelligence and
  Statistics}}, pages 1352--1362. PMLR, 2020.

\bibitem[Fort and Scherlis(2019)]{fort2019goldilocks}
Stanislav Fort and Adam Scherlis.
\newblock {The Goldilocks Zone: Towards Better Understanding of Neural Network
  Loss Landscapes}.
\newblock In \emph{{Proceedings of the AAAI Conference on Artificial
  Intelligence}}, volume~33, pages 3574--3581, 2019.

\bibitem[Srivastava et~al.(2003)Srivastava, Lee, Simoncelli, and
  Zhu]{srivastava2003advances}
Anuj Srivastava, Ann~B Lee, Eero~P Simoncelli, and S-C Zhu.
\newblock {On advances in statistical modeling of natural images}.
\newblock \emph{Journal of mathematical imaging and vision}, 18\penalty0
  (1):\penalty0 17--33, 2003.

\bibitem[Simoncelli(2009)]{simoncelli2009capturing}
Eero~P Simoncelli.
\newblock {Capturing visual image properties with probabilistic models}.
\newblock In \emph{The Essential Guide to Image Processing}, pages 205--223.
  Elsevier, 2009.

\bibitem[Pearce et~al.(2020{\natexlab{a}})Pearce, Foong, and
  Brintrup]{pearce2020structured}
Tim Pearce, Andrew~YK Foong, and Alexandra Brintrup.
\newblock {Structured Weight Priors for Convolutional Neural Networks}.
\newblock \emph{arXiv preprint arXiv:2007.14235}, 2020{\natexlab{a}}.

\bibitem[Karaletsos et~al.(2018)Karaletsos, Dayan, and
  Ghahramani]{karaletsos2018probabilistic}
Theofanis Karaletsos, Peter Dayan, and Zoubin Ghahramani.
\newblock {Probabilistic meta-representations of neural networks}.
\newblock \emph{arXiv preprint arXiv:1810.00555}, 2018.

\bibitem[Karaletsos and Bui(2020)]{karaletsos2020hierarchical}
Theofanis Karaletsos and Thang~D Bui.
\newblock {Hierarchical Gaussian Process Priors for Bayesian Neural Network
  Weights}.
\newblock \emph{Advances in Neural Information Processing Systems}, 33, 2020.

\bibitem[Atanov et~al.(2018)Atanov, Ashukha, Struminsky, Vetrov, and
  Welling]{atanov2018deep}
Andrei Atanov, Arsenii Ashukha, Kirill Struminsky, Dmitriy Vetrov, and Max
  Welling.
\newblock {The Deep Weight Prior}.
\newblock In \emph{{International Conference on Learning Representations}},
  2018.

\bibitem[Garriga-Alonso and Fortuin(2021)]{garriga2021exact}
Adri{\`a} Garriga-Alonso and Vincent Fortuin.
\newblock {Exact Langevin dynamics with stochastic gradients}.
\newblock \emph{arXiv preprint arXiv:2102.01691}, 2021.

\bibitem[Fortuin et~al.(2021{\natexlab{c}})Fortuin, Garriga-Alonso, van~der
  Wilk, and Aitchison]{fortuin2021bnnpriors}
Vincent Fortuin, Adri{\`a} Garriga-Alonso, Mark van~der Wilk, and Laurence
  Aitchison.
\newblock {BNNpriors: A library for Bayesian neural network inference with
  different prior distributions}.
\newblock \emph{Software Impacts}, page 100079, 2021{\natexlab{c}}.

\bibitem[Lippmann(1989)]{lippmann1989pattern}
Richard~P Lippmann.
\newblock {Pattern classification using neural networks}.
\newblock \emph{IEEE communications magazine}, 27\penalty0 (11):\penalty0
  47--50, 1989.

\bibitem[Coker et~al.(2019)Coker, Pradier, and Doshi-Velez]{coker2019towards}
Beau Coker, Melanie~F Pradier, and Finale Doshi-Velez.
\newblock {Towards Expressive Priors for Bayesian Neural Networks: Poisson
  Process Radial Basis Function Networks}.
\newblock \emph{arXiv preprint arXiv:1912.05779}, 2019.

\bibitem[Brea et~al.(2019)Brea, Simsek, Illing, and Gerstner]{brea2019weight}
Johanni Brea, Berfin Simsek, Bernd Illing, and Wulfram Gerstner.
\newblock {Weight-space symmetry in deep networks gives rise to permutation
  saddles, connected by equal-loss valleys across the loss landscape}.
\newblock \emph{arXiv preprint arXiv:1907.02911}, 2019.

\bibitem[Fort and Jastrzebski(2019)]{fort2019large}
Stanislav Fort and Stanislaw Jastrzebski.
\newblock {Large scale structure of neural network loss landscapes}.
\newblock \emph{arXiv preprint arXiv:1906.04724}, 2019.

\bibitem[Sun et~al.(2018{\natexlab{a}})Sun, Zhang, Shi, and
  Grosse]{sun2018functional}
Shengyang Sun, Guodong Zhang, Jiaxin Shi, and Roger Grosse.
\newblock {Functional Variational Bayesian Neural Networks}.
\newblock In \emph{{International Conference on Learning Representations}},
  2018{\natexlab{a}}.

\bibitem[Flam-Shepherd et~al.(2017)Flam-Shepherd, Requeima, and
  Duvenaud]{flam2017mapping}
Daniel Flam-Shepherd, James Requeima, and David Duvenaud.
\newblock {Mapping Gaussian process priors to Bayesian neural networks}.
\newblock In \emph{{NeurIPS} {B}ayesian deep learning workshop}, 2017.

\bibitem[Burt et~al.(2020)Burt, Ober, Garriga-Alonso, and van~der
  Wilk]{burt2020understanding}
David~R Burt, Sebastian~W Ober, Adri{\`a} Garriga-Alonso, and Mark van~der
  Wilk.
\newblock {Understanding Variational Inference in Function-Space}.
\newblock \emph{arXiv preprint arXiv:2011.09421}, 2020.

\bibitem[Tran et~al.(2020)Tran, Rossi, Milios, and Filippone]{tran2020all}
Ba-Hien Tran, Simone Rossi, Dimitrios Milios, and Maurizio Filippone.
\newblock {All You Need is a Good Functional Prior for Bayesian Deep Learning}.
\newblock \emph{arXiv preprint arXiv:2011.12829}, 2020.

\bibitem[Ha et~al.(2016)Ha, Dai, and Le]{ha2016hypernetworks}
David Ha, Andrew Dai, and Quoc~V Le.
\newblock {Hypernetworks}.
\newblock \emph{arXiv preprint arXiv:1609.09106}, 2016.

\bibitem[Krueger et~al.(2017)Krueger, Huang, Islam, Turner, Lacoste, and
  Courville]{krueger2017bayesian}
David Krueger, Chin-Wei Huang, Riashat Islam, Ryan Turner, Alexandre Lacoste,
  and Aaron Courville.
\newblock {Bayesian hypernetworks}.
\newblock \emph{arXiv preprint arXiv:1710.04759}, 2017.

\bibitem[Flam-Shepherd et~al.(2018)Flam-Shepherd, Requeima, and
  Duvenaud]{flam2018characterizing}
Daniel Flam-Shepherd, James Requeima, and David Duvenaud.
\newblock {Characterizing and Warping the Function Space of Bayesian Neural
  Networks}.
\newblock In \emph{{NeurIPS Workshop on Bayesian Deep Learning}}, 2018.

\bibitem[Candes(1998)]{candes1998ridgelets}
Emmanuel~Jean Candes.
\newblock {Ridgelets: Theory and application}.
\newblock \emph{Ph. D. dissertation, Dept. of Statistics, Stanford Univ.},
  1998.

\bibitem[Matsubara et~al.(2020)Matsubara, Oates, and
  Briol]{matsubara2020ridgelet}
Takuo Matsubara, Chris~J Oates, and Fran{\c{c}}ois-Xavier Briol.
\newblock {The Ridgelet Prior: A Covariance Function Approach to Prior
  Specification for Bayesian Neural Networks}.
\newblock \emph{arXiv preprint arXiv:2010.08488}, 2020.

\bibitem[Ma et~al.(2019{\natexlab{b}})Ma, Li, and
  Hern{\'a}ndez-Lobato]{ma2019variational}
Chao Ma, Yingzhen Li, and Jos{\'e}~Miguel Hern{\'a}ndez-Lobato.
\newblock {Variational implicit processes}.
\newblock In \emph{{International Conference on Machine Learning}}, pages
  4222--4233. PMLR, 2019{\natexlab{b}}.

\bibitem[Pearce et~al.(2020{\natexlab{b}})Pearce, Tsuchida, Zaki, Brintrup, and
  Neely]{pearce2020expressive}
Tim Pearce, Russell Tsuchida, Mohamed Zaki, Alexandra Brintrup, and Andy Neely.
\newblock {Expressive priors in Bayesian neural networks: Kernel combinations
  and periodic functions}.
\newblock In \emph{Uncertainty in Artificial Intelligence}, pages 134--144.
  PMLR, 2020{\natexlab{b}}.

\bibitem[Yang et~al.(2019)Yang, Lorch, Graule, Srinivasan, Suresh, Yao,
  Pradier, and Doshi-Velez]{yang2019output}
Wanqian Yang, Lars Lorch, Moritz~A Graule, Srivatsan Srinivasan, Anirudh
  Suresh, Jiayu Yao, Melanie~F Pradier, and Finale Doshi-Velez.
\newblock {Output-constrained Bayesian neural networks}.
\newblock \emph{arXiv preprint arXiv:1905.06287}, 2019.

\bibitem[Hafner et~al.(2020)Hafner, Tran, Lillicrap, Irpan, and
  Davidson]{hafner2020noise}
Danijar Hafner, Dustin Tran, Timothy Lillicrap, Alex Irpan, and James Davidson.
\newblock {Noise contrastive priors for functional uncertainty}.
\newblock In \emph{{Uncertainty in Artificial Intelligence}}, pages 905--914.
  PMLR, 2020.

\bibitem[Nalisnick et~al.(2021)Nalisnick, Gordon, and
  Hern{\'a}ndez-Lobato]{nalisnick2021predictive}
Eric Nalisnick, Jonathan Gordon, and Jos{\'e}~Miguel Hern{\'a}ndez-Lobato.
\newblock {Predictive Complexity Priors}.
\newblock In \emph{{International Conference on Artificial Intelligence and
  Statistics}}, pages 694--702. PMLR, 2021.

\bibitem[Efron and Tibshirani(1994)]{efron1994introduction}
Bradley Efron and Robert~J Tibshirani.
\newblock \emph{{An introduction to the bootstrap}}.
\newblock CRC press, 1994.

\bibitem[Lakshminarayanan et~al.(2017)Lakshminarayanan, Pritzel, and
  Blundell]{lakshminarayanan2017simple}
Balaji Lakshminarayanan, Alexander Pritzel, and Charles Blundell.
\newblock {Simple and scalable predictive uncertainty estimation using deep
  ensembles}.
\newblock In \emph{{Proceedings of the 31st International Conference on Neural
  Information Processing Systems}}, pages 6405--6416, 2017.

\bibitem[Matthews et~al.(2017)Matthews, Hron, Turner, and
  Ghahramani]{matthews2017sample}
Alexander G de~G Matthews, Jiri Hron, Richard~E Turner, and Zoubin Ghahramani.
\newblock {Sample-then-optimize posterior sampling for bayesian linear models}.
\newblock In \emph{NeurIPS Workshop on Advances in Approximate Bayesian
  Inference}, 2017.

\bibitem[Lyle et~al.(2020)Lyle, Schut, Ru, Gal, and van~der
  Wilk]{lyle2020bayesian}
Clare Lyle, Lisa Schut, Robin Ru, Yarin Gal, and Mark van~der Wilk.
\newblock {A Bayesian Perspective on Training Speed and Model Selection}.
\newblock \emph{Advances in Neural Information Processing Systems}, 33, 2020.

\bibitem[Wenzel et~al.(2020{\natexlab{b}})Wenzel, Snoek, Tran, and
  Jenatton]{wenzel2020hyper}
Florian Wenzel, Jasper Snoek, Dustin Tran, and Rodolphe Jenatton.
\newblock {Hyperparameter Ensembles for Robustness and Uncertainty
  Quantification}.
\newblock In \emph{Advances in Neural Information Processing Systems},
  2020{\natexlab{b}}.

\bibitem[Wen et~al.(2019)Wen, Tran, and Ba]{wen2019batchensemble}
Yeming Wen, Dustin Tran, and Jimmy Ba.
\newblock {BatchEnsemble: an Alternative Approach to Efficient Ensemble and
  Lifelong Learning}.
\newblock In \emph{{International Conference on Learning Representations}},
  2019.

\bibitem[Rahaman and Thiery(2020)]{rahaman2020uncertainty}
Rahul Rahaman and Alexandre~H Thiery.
\newblock {Uncertainty quantification and deep ensembles}.
\newblock \emph{arXiv preprint arXiv:2007.08792}, 2020.

\bibitem[Yao et~al.(2019)Yao, Pan, Ghosh, and Doshi-Velez]{yao2019quality}
Jiayu Yao, Weiwei Pan, Soumya Ghosh, and Finale Doshi-Velez.
\newblock {Quality of uncertainty quantification for Bayesian neural network
  inference}.
\newblock \emph{arXiv preprint arXiv:1906.09686}, 2019.

\bibitem[Osband et~al.(2018)Osband, Aslanides, and
  Cassirer]{osband2018randomized}
Ian Osband, John Aslanides, and Albin Cassirer.
\newblock {Randomized prior functions for deep reinforcement learning}.
\newblock In \emph{{Proceedings of the 32nd International Conference on Neural
  Information Processing Systems}}, pages 8626--8638, 2018.

\bibitem[Osband et~al.(2019)Osband, Van~Roy, Russo, and Wen]{osband2019deep}
Ian Osband, Benjamin Van~Roy, Daniel~J Russo, and Zheng Wen.
\newblock {Deep Exploration via Randomized Value Functions.}
\newblock \emph{Journal of Machine Learning Research}, 20\penalty0
  (124):\penalty0 1--62, 2019.

\bibitem[Ciosek et~al.(2020)Ciosek, Fortuin, Tomioka, Hofmann, and
  Turner]{ciosek2019conservative}
Kamil Ciosek, Vincent Fortuin, Ryota Tomioka, Katja Hofmann, and Richard
  Turner.
\newblock {Conservative uncertainty estimation by fitting prior networks}.
\newblock In \emph{{International Conference on Learning Representations}},
  2020.

\bibitem[He et~al.(2020)He, Lakshminarayanan, and Teh]{he2020bayesian}
Bobby He, Balaji Lakshminarayanan, and Yee~Whye Teh.
\newblock {Bayesian deep ensembles via the neural tangent kernel}.
\newblock \emph{arXiv preprint arXiv:2007.05864}, 2020.

\bibitem[Liu and Wang(2016)]{liu2016stein}
Qiang Liu and Dilin Wang.
\newblock {Stein variational Gradient descent: a general purpose Bayesian
  inference algorithm}.
\newblock In \emph{{Proceedings of the 30th International Conference on Neural
  Information Processing Systems}}, pages 2378--2386, 2016.

\bibitem[Liu(2017)]{liu2017stein}
Qiang Liu.
\newblock {Stein variational gradient descent as gradient flow}.
\newblock In \emph{{Proceedings of the 31st International Conference on Neural
  Information Processing Systems}}, pages 3118--3126, 2017.

\bibitem[Korba et~al.(2020)Korba, Salim, Arbel, Luise, and
  Gretton]{korba2020non}
Anna Korba, Adil Salim, Michael Arbel, Giulia Luise, and Arthur Gretton.
\newblock {A non-asymptotic analysis for Stein variational gradient descent}.
\newblock \emph{arXiv preprint arXiv:2006.09797}, 2020.

\bibitem[Hu et~al.(2019)Hu, Szerlip, Karaletsos, and Singh]{hu2019applying}
Xinyu Hu, Paul Szerlip, Theofanis Karaletsos, and Rohit Singh.
\newblock {Applying SVGD to Bayesian Neural Networks for Cyclical Time-Series
  Prediction and Inference}.
\newblock \emph{arXiv preprint arXiv:1901.05906}, 2019.

\bibitem[D'Angelo et~al.(2021)D'Angelo, Fortuin, and Wenzel]{d2021stein}
Francesco D'Angelo, Vincent Fortuin, and Florian Wenzel.
\newblock {On Stein Variational Neural Network Ensembles}.
\newblock \emph{arXiv preprint arXiv:2106.10760}, 2021.

\bibitem[Wang et~al.(2018)Wang, Ren, Zhu, and Zhang]{wang2018function}
Ziyu Wang, Tongzheng Ren, Jun Zhu, and Bo~Zhang.
\newblock {Function Space Particle Optimization for Bayesian Neural Networks}.
\newblock In \emph{{International Conference on Learning Representations}},
  2018.

\bibitem[D'Angelo and Fortuin(2021)]{d2021repulsive}
Francesco D'Angelo and Vincent Fortuin.
\newblock {Repulsive Deep Ensembles are Bayesian}.
\newblock In \emph{{Advances in Neural Information Processing Systems}}, 2021.

\bibitem[Rasmussen and Ghahramani(2001)]{rasmussen2001occam}
Carl~E Rasmussen and Zoubin Ghahramani.
\newblock {Occam's razor}.
\newblock \emph{Advances in Neural Information Processing Systems}, pages
  294--300, 2001.

\bibitem[Schmidhuber(1987)]{schmidhuber1987evolutionary}
J{\"u}rgen Schmidhuber.
\newblock \emph{{Evolutionary principles in self-referential learning, or on
  learning how to learn: the meta-meta-... hook}}.
\newblock PhD thesis, Technische Universit{\"a}t M{\"u}nchen, 1987.

\bibitem[Thrun and Pratt(1998)]{thrun1998learning}
Sebastian Thrun and Lorien Pratt.
\newblock {Learning to learn: Introduction and overview}.
\newblock In \emph{{Learning to learn}}, pages 3--17. Springer, 1998.

\bibitem[Baxter(2000)]{baxter2000model}
Jonathan Baxter.
\newblock {A model of inductive bias learning}.
\newblock \emph{Journal of artificial intelligence research}, 12:\penalty0
  149--198, 2000.

\bibitem[Heskes(1998)]{heskes1998solving}
Tom Heskes.
\newblock {Solving a Huge Number of Similar Tasks: A Combination of Multi-Task
  Learning and a Hierarchical Bayesian Approach}.
\newblock In \emph{{Proceedings of the Fifteenth International Conference on
  Machine Learning}}, pages 233--241, 1998.

\bibitem[Tenenbaum(1999)]{tenenbaum1999bayesian}
Joshua~B Tenenbaum.
\newblock \emph{{A Bayesian Framework for Concept Learning}}.
\newblock PhD thesis, Citeseer, 1999.

\bibitem[Fei-Fei et~al.(2003)Fei-Fei, Fergus, and Perona]{fei2003bayesian}
Li~Fei-Fei, Rob Fergus, and Pietro Perona.
\newblock {A Bayesian approach to unsupervised one-shot learning of object
  categories}.
\newblock In \emph{{Proceedings Ninth IEEE International Conference on Computer
  Vision}}, pages 1134--1141. IEEE, 2003.

\bibitem[Lawrence and Platt(2004)]{lawrence2004learning}
Neil~D Lawrence and John~C Platt.
\newblock {Learning to learn with the informative vector machine}.
\newblock In \emph{{Proceedings of the twenty-first international conference on
  Machine learning}}, page~65, 2004.

\bibitem[Grant et~al.(2018)Grant, Finn, Levine, Darrell, and
  Griffiths]{grant2018recasting}
Erin Grant, Chelsea Finn, Sergey Levine, Trevor Darrell, and Thomas Griffiths.
\newblock {Recasting Gradient-Based Meta-Learning as Hierarchical Bayes}.
\newblock In \emph{{International Conference on Learning Representations}},
  2018.

\bibitem[Yoon et~al.(2018)Yoon, Kim, Dia, Kim, Bengio, and
  Ahn]{yoon2018bayesian}
Jaesik Yoon, Taesup Kim, Ousmane Dia, Sungwoong Kim, Yoshua Bengio, and Sungjin
  Ahn.
\newblock {Bayesian model-agnostic meta-learning}.
\newblock In \emph{{Proceedings of the 32nd International Conference on Neural
  Information Processing Systems}}, pages 7343--7353, 2018.

\bibitem[Finn et~al.(2018)Finn, Xu, and Levine]{finn2018probabilistic}
Chelsea Finn, Kelvin Xu, and Sergey Levine.
\newblock {Probabilistic model-agnostic meta-learning}.
\newblock In \emph{{Proceedings of the 32nd International Conference on Neural
  Information Processing Systems}}, pages 9537--9548, 2018.

\bibitem[Wilson and Adams(2013)]{wilson2013gaussian}
Andrew Wilson and Ryan Adams.
\newblock {Gaussian process kernels for pattern discovery and extrapolation}.
\newblock In \emph{{International conference on machine learning}}, pages
  1067--1075. PMLR, 2013.

\bibitem[Duvenaud et~al.(2013)Duvenaud, Lloyd, Grosse, Tenenbaum, and
  Zoubin]{duvenaud2013structure}
David Duvenaud, James Lloyd, Roger Grosse, Joshua Tenenbaum, and Ghahramani
  Zoubin.
\newblock {Structure discovery in nonparametric regression through
  compositional kernel search}.
\newblock In \emph{{International Conference on Machine Learning}}, pages
  1166--1174. PMLR, 2013.

\bibitem[Lloyd et~al.(2014)Lloyd, Duvenaud, Grosse, Tenenbaum, and
  Ghahramani]{lloyd2014automatic}
James Lloyd, David Duvenaud, Roger Grosse, Joshua Tenenbaum, and Zoubin
  Ghahramani.
\newblock {Automatic construction and natural-language description of
  nonparametric regression models}.
\newblock In \emph{{Proceedings of the AAAI Conference on Artificial
  Intelligence}}, volume~28, 2014.

\bibitem[Kim and Teh(2018)]{kim2018scaling}
Hyunjik Kim and Yee~Whye Teh.
\newblock {Scaling up the Automatic Statistician: Scalable structure discovery
  using Gaussian processes}.
\newblock In \emph{{International Conference on Artificial Intelligence and
  Statistics}}, pages 575--584. PMLR, 2018.

\bibitem[Sun et~al.(2018{\natexlab{b}})Sun, Zhang, Wang, Zeng, Li, and
  Grosse]{sun2018differentiable}
Shengyang Sun, Guodong Zhang, Chaoqi Wang, Wenyuan Zeng, Jiaman Li, and Roger
  Grosse.
\newblock {Differentiable compositional kernel learning for Gaussian
  processes}.
\newblock In \emph{{International Conference on Machine Learning}}, pages
  4828--4837. PMLR, 2018{\natexlab{b}}.

\bibitem[Liu et~al.(2020)Liu, Lin, Padhy, Tran, Bedrax-Weiss, and
  Lakshminarayanan]{liu2020simple}
Jeremiah~Zhe Liu, Zi~Lin, Shreyas Padhy, Dustin Tran, Tania Bedrax-Weiss, and
  Balaji Lakshminarayanan.
\newblock {Simple and principled uncertainty estimation with deterministic deep
  learning via distance awareness}.
\newblock \emph{arXiv preprint arXiv:2006.10108}, 2020.

\bibitem[Fortuin et~al.(2021{\natexlab{d}})Fortuin, Collier, Wenzel, Allingham,
  Liu, Tran, Lakshminarayanan, Berent, Jenatton, and
  Kokiopoulou]{fortuin2021deep}
Vincent Fortuin, Mark Collier, Florian Wenzel, James Allingham, Jeremiah Liu,
  Dustin Tran, Balaji Lakshminarayanan, Jesse Berent, Rodolphe Jenatton, and
  Effrosyni Kokiopoulou.
\newblock {Deep Classifiers with Label Noise Modeling and Distance Awareness}.
\newblock \emph{arXiv preprint arXiv:2110.02609}, 2021{\natexlab{d}}.

\bibitem[Ober et~al.(2021)Ober, Rasmussen, and van~der Wilk]{ober2021promises}
Sebastian~W Ober, Carl~E Rasmussen, and Mark van~der Wilk.
\newblock {The Promises and Pitfalls of Deep Kernel Learning}.
\newblock \emph{arXiv preprint arXiv:2102.12108}, 2021.

\bibitem[van~der Wilk et~al.(2018)van~der Wilk, Bauer, John, and
  Hensman]{vdw2018inv}
Mark van~der Wilk, Matthias Bauer, ST~John, and James Hensman.
\newblock {Learning Invariances using the Marginal Likelihood}.
\newblock In \emph{Advances in Neural Information Processing Systems},
  volume~31, pages 9938--9948, 2018.

\bibitem[Patacchiola et~al.(2020)Patacchiola, Turner, Crowley, O'Boyle, and
  Storkey]{patacchiola2020bayesian}
Massimiliano Patacchiola, Jack Turner, Elliot~J Crowley, Michael O'Boyle, and
  Amos~J Storkey.
\newblock {Bayesian Meta-Learning for the Few-Shot Setting via Deep Kernels}.
\newblock \emph{Advances in Neural Information Processing Systems}, 33, 2020.

\bibitem[Qin et~al.(2018)Qin, Zhang, Zhao, Wang, Shi, Qi, Shi, and
  Lei]{qin2018rethink}
Yunxiao Qin, Weiguo Zhang, Chenxu Zhao, Zezheng Wang, Hailin Shi, Guojun Qi,
  Jingping Shi, and Zhen Lei.
\newblock {Rethink and redesign meta learning}.
\newblock \emph{arXiv preprint arXiv:1812.04955}, 2018.

\bibitem[Yin et~al.(2019)Yin, Tucker, Zhou, Levine, and Finn]{yin2019meta}
Mingzhang Yin, George Tucker, Mingyuan Zhou, Sergey Levine, and Chelsea Finn.
\newblock {Meta-Learning without Memorization}.
\newblock In \emph{{International Conference on Learning Representations}},
  2019.

\bibitem[Rothfuss et~al.(2021)Rothfuss, Fortuin, Josifoski, and
  Krause]{rothfuss2020pacoh}
Jonas Rothfuss, Vincent Fortuin, Martin Josifoski, and Andreas Krause.
\newblock {PACOH: Bayes-Optimal Meta-Learning with PAC-Guarantees}.
\newblock In \emph{{International Conference on Machine Learning}}, 2021.

\bibitem[Hoffman and Johnson(2016)]{hoffman2016elbo}
Matthew~D Hoffman and Matthew~J Johnson.
\newblock {ELBO Surgery: Yet another way to carve up the evidence lower bound}.
\newblock In \emph{Advances in Approximate Bayesian Inference}, 2016.

\bibitem[Tomczak and Welling(2018)]{tomczak2018vae}
Jakub Tomczak and Max Welling.
\newblock {VAE with a VampPrior}.
\newblock In \emph{{International Conference on Artificial Intelligence and
  Statistics}}, pages 1214--1223. PMLR, 2018.

\bibitem[Dalal and Hall(1983)]{dalal1983approximating}
SR~Dalal and WJ~Hall.
\newblock {Approximating priors by mixtures of natural conjugate priors}.
\newblock \emph{Journal of the Royal Statistical Society: Series B
  (Methodological)}, 45\penalty0 (2):\penalty0 278--286, 1983.

\bibitem[Guo et~al.(2020)Guo, Zhou, Chen, Ying, Zhang, and
  Zhou]{guo2020variational}
Chunsheng Guo, Jialuo Zhou, Huahua Chen, Na~Ying, Jianwu Zhang, and Di~Zhou.
\newblock {Variational autoencoder with optimizing Gaussian mixture model
  priors}.
\newblock \emph{IEEE Access}, 8:\penalty0 43992--44005, 2020.

\bibitem[Botros and Tomczak(2018)]{botros2018hierarchical}
Philip Botros and Jakub~M Tomczak.
\newblock {Hierarchical VampPrior variational fair auto-encoder}.
\newblock \emph{arXiv preprint arXiv:1806.09918}, 2018.

\bibitem[Gulrajani et~al.(2017)Gulrajani, Kumar, Ahmed, Taiga, Visin, Vazquez,
  and Courville]{gulrajani2016pixelvae}
Ishaan Gulrajani, Kundan Kumar, Faruk Ahmed, Adrien~Ali Taiga, Francesco Visin,
  David Vazquez, and Aaron Courville.
\newblock {Pixelvae: A latent variable model for natural images}.
\newblock \emph{International Conference on Learning Representations}, 2017.

\bibitem[Bornschein et~al.(2017)Bornschein, Mnih, Zoran, and
  Rezende]{bornschein2017variational}
J{\"o}rg Bornschein, Andriy Mnih, Daniel Zoran, and Danilo~J Rezende.
\newblock {Variational memory addressing in generative models}.
\newblock In \emph{{Proceedings of the 31st International Conference on Neural
  Information Processing Systems}}, pages 3923--3932, 2017.

\bibitem[Rezende and Mohamed(2015)]{rezende2015variational}
Danilo Rezende and Shakir Mohamed.
\newblock {Variational inference with normalizing flows}.
\newblock In \emph{{International Conference on Machine Learning}}, pages
  1530--1538. PMLR, 2015.

\bibitem[Dinh et~al.(2016)Dinh, Sohl-Dickstein, and Bengio]{dinh2016density}
Laurent Dinh, Jascha Sohl-Dickstein, and Samy Bengio.
\newblock {Density estimation using real nvp}.
\newblock \emph{arXiv preprint arXiv:1605.08803}, 2016.

\bibitem[Huang et~al.(2017)Huang, Touati, Dinh, Drozdzal, Havaei, Charlin, and
  Courville]{huang2017learnable}
Chin-Wei Huang, Ahmed Touati, Laurent Dinh, Michal Drozdzal, Mohammad Havaei,
  Laurent Charlin, and Aaron Courville.
\newblock {Learnable explicit density for continuous latent space and
  variational inference}.
\newblock \emph{arXiv preprint arXiv:1710.02248}, 2017.

\bibitem[Kingma et~al.(2016)Kingma, Salimans, Jozefowicz, Chen, Sutskever, and
  Welling]{kingma2016improved}
Diederik~P Kingma, Tim Salimans, Rafal Jozefowicz, Xi~Chen, Ilya Sutskever, and
  Max Welling.
\newblock {Improved variational inference with inverse autoregressive flow}.
\newblock In \emph{{Proceedings of the 30th International Conference on Neural
  Information Processing Systems}}, pages 4743--4751, 2016.

\bibitem[Chen et~al.(2016)Chen, Kingma, Salimans, Duan, Dhariwal, Schulman,
  Sutskever, and Abbeel]{chen2016variational}
Xi~Chen, Diederik~P Kingma, Tim Salimans, Yan Duan, Prafulla Dhariwal, John
  Schulman, Ilya Sutskever, and Pieter Abbeel.
\newblock {Variational lossy autoencoder}.
\newblock \emph{arXiv preprint arXiv:1611.02731}, 2016.

\bibitem[Bauer and Mnih(2019)]{bauer2019resampled}
Matthias Bauer and Andriy Mnih.
\newblock {Resampled priors for variational autoencoders}.
\newblock In \emph{{The 22nd International Conference on Artificial
  Intelligence and Statistics}}, pages 66--75. PMLR, 2019.

\bibitem[Pang et~al.(2020)Pang, Han, Nijkamp, Zhu, and Wu]{pang2020learning}
Bo~Pang, Tian Han, Erik Nijkamp, Song-Chun Zhu, and Ying~Nian Wu.
\newblock {Learning Latent Space Energy-Based Prior Model}.
\newblock \emph{Advances in Neural Information Processing Systems}, 33, 2020.

\bibitem[Aneja et~al.(2020)Aneja, Schwing, Kautz, and Vahdat]{aneja2020ncp}
Jyoti Aneja, Alexander Schwing, Jan Kautz, and Arash Vahdat.
\newblock {NCP-VAE: Variational Autoencoders with Noise Contrastive Priors}.
\newblock \emph{arXiv preprint arXiv:2010.02917}, 2020.

\bibitem[Chen et~al.(2020)Chen, Liu, Cai, Xu, and Pei]{chen2020vaepp}
Wenxiao Chen, Wenda Liu, Zhenting Cai, Haowen Xu, and Dan Pei.
\newblock {VAEPP: Variational Autoencoder with a Pull-Back Prior}.
\newblock In \emph{{International Conference on Neural Information
  Processing}}, pages 366--379. Springer, 2020.

\bibitem[Goodfellow et~al.(2014)Goodfellow, Pouget-Abadie, Mirza, Xu,
  Warde-Farley, Ozair, Courville, and Bengio]{goodfellow2014generative}
Ian~J Goodfellow, Jean Pouget-Abadie, Mehdi Mirza, Bing Xu, David Warde-Farley,
  Sherjil Ozair, Aaron Courville, and Yoshua Bengio.
\newblock {Generative adversarial nets}.
\newblock In \emph{{Proceedings of the 27th International Conference on Neural
  Information Processing Systems-Volume 2}}, pages 2672--2680, 2014.

\bibitem[Immer et~al.(2021{\natexlab{b}})Immer, Bauer, Fortuin, R{\"a}tsch, and
  Khan]{immer2021scalable}
Alexander Immer, Matthias Bauer, Vincent Fortuin, Gunnar R{\"a}tsch, and
  Mohammad~Emtiyaz Khan.
\newblock {Scalable Marginal Likelihood Estimation for Model Selection in Deep
  Learning}.
\newblock In \emph{{International Conference on Machine Learning}},
  2021{\natexlab{b}}.

\bibitem[Nalisnick and Smyth(2018)]{nalisnick2018learning}
Eric Nalisnick and Padhraic Smyth.
\newblock {Learning priors for invariance}.
\newblock In \emph{{International Conference on Artificial Intelligence and
  Statistics}}, pages 366--375. PMLR, 2018.

\end{thebibliography}

\end{document}